\documentclass[conference]{IEEEtran}
\usepackage{times}

\usepackage[utf8]{inputenc}
\usepackage[T1]{fontenc}
\usepackage{textcomp}       
\usepackage{gensymb}        
\usepackage{microtype}      

\usepackage{amsmath,amssymb,amsfonts,bbm}
\usepackage{amsthm}
\usepackage{nicefrac}       
\usepackage{siunitx}        

\usepackage[dvipsnames,svgnames,table]{xcolor} 
\usepackage{graphicx}
\usepackage{subcaption}
\usepackage{booktabs}       
\usepackage{multirow}
\usepackage{makecell}
\usepackage{array}          
\usepackage{tabularx}
\usepackage{float}
\usepackage{wrapfig}
\usepackage{multicol}
\usepackage[ruled, vlined, linesnumbered]{algorithm2e}

\SetCommentSty{mycommfont}
\SetKwComment{Comment}{$\triangleright$\ }{}

\usepackage[numbers]{natbib}
\usepackage{url}

\usepackage[bookmarks=true]{hyperref}
\usepackage[capitalize,noabbrev]{cleveref}

\crefname{equation}{Eq.}{Eqs.}

\hypersetup{
    colorlinks=true,
    linkcolor=black,
    urlcolor=brown,
    citecolor=black,
}

\usepackage{xcolor}

\definecolor{cFMBC}{HTML}{38AABF}
\definecolor{cMFBC}{HTML}{3BA3EC}
\definecolor{cQCFQL}{HTML}{38AABF}
\definecolor{cQCMFQL}{HTML}{3BA3EC}
\definecolor{cFMLQL}{HTML}{BB9832}
\definecolor{cDSRL}{HTML}{E68332}
\definecolor{cLPS}{HTML}{F77189}
\definecolor{cLPSD}{HTML}{50B131}
\definecolor{cCFGRL}{HTML}{A48CF4}
\definecolor{cDataset}{HTML}{808080} 

\begin{document}

\title{Latent Policy Steering through One-Step Flow Policies}

\author{
  Hokyun Im$^1$ \quad Andrey Kolobov$^2$ \quad Jianlong Fu$^2$ \quad Youngwoon Lee$^1$ \\
  $^1$Department of Artificial Intelligence, Yonsei University \quad $^2$ Microsoft Research\\
\texttt{\url{https://jellyho.github.io/LPS/}}
}

\maketitle

\begin{abstract}
Offline reinforcement learning (RL) allows robots to learn from offline datasets without risky exploration. Yet, offline RL's performance often hinges on a brittle trade-off between (1) return maximization, which can push policies outside the dataset support, and (2) behavioral constraints, which typically require sensitive hyperparameter tuning. Latent steering offers a structural way to stay within the dataset support during RL, but existing offline adaptations commonly approximate action values using latent-space critics learned via indirect distillation, which can lose information and hinder convergence. We propose Latent Policy Steering (LPS), which enables high-fidelity latent policy improvement by backpropagating original-action-space Q-gradients through a differentiable one-step MeanFlow policy to update a latent-action-space actor. By eliminating proxy latent critics, LPS allows an original-action-space critic to guide end-to-end latent-space optimization, while the one-step MeanFlow policy serves as a behavior-constrained generative prior. This decoupling yields a robust method that works out-of-the-box with minimal tuning. Across OGBench and real-world robotic tasks, LPS achieves state-of-the-art performance and consistently outperforms behavioral cloning and strong latent steering baselines.
\end{abstract}

\IEEEpeerreviewmaketitle

\section{Introduction}
\label{sec:introduction}

Offline reinforcement learning (RL) promises to enable robots to acquire complex behaviors from large-scale, pre-collected datasets without costly and dangerous real-world interaction. Despite recent progress in offline RL~\citep{DBLP:conf/iclr/WangHZ23,wang2025onestepgenerativepoliciesqlearning,DBLP:conf/icml/ParkLL25,li2025qc} and impressive results in simulation~\citep{DBLP:conf/iclr/ParkFEL25}, reliably transferring these methods to real-world robotics remains challenging.

Most state-of-the-art offline RL algorithms follow the TD3+BC~\citep{DBLP:conf/nips/FujimotoG21} paradigm and its generative variants with more expressive regularization~\citep{DBLP:conf/icml/ParkLL25,li2025qc}. These approaches aim to maximize return while constraining the learned policy to the dataset support by adding a regularization term, weighted by a hyperparameter $\alpha$. In practice, this formulation introduces a delicate trade-off: weak regularization leads to out-of-distribution actions and extrapolation error, while excessive regularization reduces offline RL to behavioral cloning. The best $\alpha$ is highly sensitive to reward scale, dataset diversity, and model capacity, making extensive hyperparameter sweeps feasible in simulation but prohibitively expensive and risky with real-world robots. This sensitivity limits the practicality and scalability of offline RL in real-world deployment.

\begin{figure}[ht]
    \vspace{-0.5em}
    \centering
    \includegraphics[width=\linewidth]{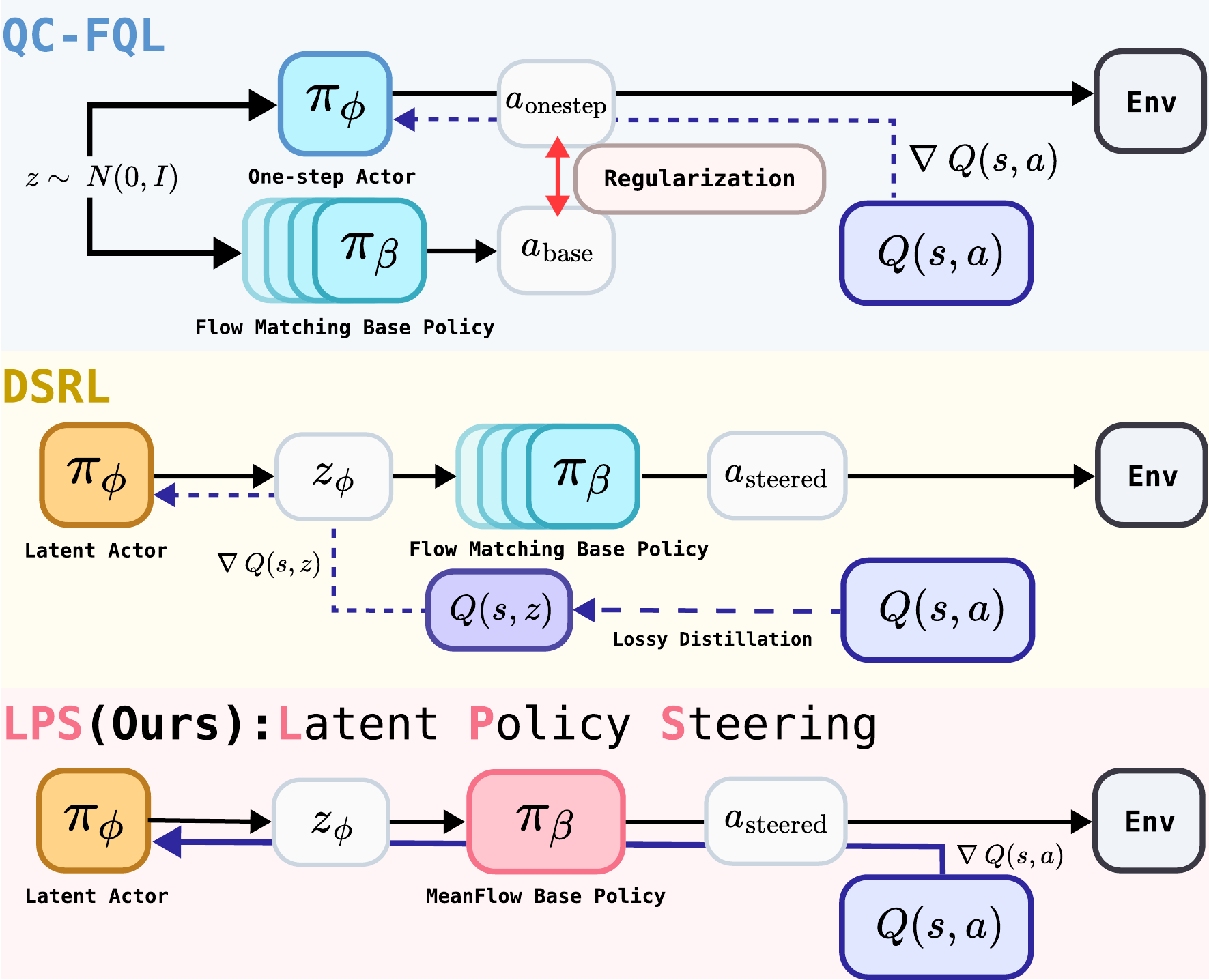}
    \caption{\textbf{Comparison of policy extraction paradigms.} 
    \textit{(Top)} \textbf{QC-FQL} constrains the policy via an explicit regularizer, creating a trade-off between reward maximization and behavioral regularization. 
    \textit{(Middle)} \textbf{DSRL} resolves this trade-off via latent steering, but requires learning a latent-space critic $Q(s,z)$ via distillation in the offline RL setting.
    \textit{(Bottom)} \textbf{LPS (Ours)} achieves robust, tuning-free optimization by backpropagating action-space critic gradients $\nabla_a Q(s,a)$ through a differentiable one-step generative policy.}
\label{fig:concept_comparison}
    \label{fig:motivation}
\end{figure}

This raises a fundamental question: can we enforce behavioral constraints safely and effectively \emph{without} relying on sensitive hyperparameter tuning? Prior work explores structural constraints via latent action models, such as VAEs~\citep{DBLP:conf/corl/ZhouBH20}, or by learning skill priors~\citep{DBLP:conf/corl/PertschLL20}. However, these methods often still require task-specific tuning or additional online interaction. In this work, we draw inspiration from recent advances in online fine-tuning of robot policies. Instead of directly updating the parameters of a generative base policy~\citep{DBLP:conf/iclr/WangHZ23,wang2025onestepgenerativepoliciesqlearning} or distilling it into a simplified one-step actor~\citep{DBLP:conf/icml/ParkLL25}, methods such as DSRL~\citep{wagenmaker2025steering} improve behavior by \emph{steering} the generation process through its latent variables. Optimizing the latent input w.r.t. a critic while keeping the pre-trained generative model fixed naturally confines the resulting policy to the data manifold, providing a form of structural regularization without an explicit regularization weight.

However, adapting this latent steering paradigm to the fully offline setting poses a key challenge. Offline datasets provide supervision for an action-space critic $Q(s,a)$, but not for a value function defined over the \emph{latent space}. DSRL addresses this mismatch with noise aliasing, distilling action-space values into an approximate latent-space critic. This additional distillation step can be lossy and may fail to capture the high-frequency details of the true value landscape, limiting the quality of offline policy improvement. As a result, such methods are often used primarily as initializations for subsequent online fine-tuning rather than as standalone offline RL solutions.

To overcome these limitations, we introduce \textbf{Latent Policy Steering (LPS)}, a framework that combines the safety of latent steering with direct value-based improvement. LPS leverages \textbf{MeanFlow}~\citep{geng2025meanflowsonestepgenerative}, a differentiable one-step generative model as our base policy, enabling efficient and stable gradient flow from the action space back to the latent space. Unlike DSRL, LPS \emph{directly} optimizes the latent actor using gradients from an action-space critic, bypassing the need for proxy latent critics while preserving the tuning-free structural constraints imposed by the generative prior (\Cref{fig:motivation}). This decoupling allows the agent to focus on policy improvement without tuning an explicit behavioral regularization weight, resulting in an out-of-the-box method that consistently matches or surpasses behavioral cloning (BC). We evaluate LPS on standard offline RL benchmarks, confirming its robust, tuning-free nature, and demonstrate strong real-world performance on robotic manipulation tasks, where it reliably improves beyond BC.

Our contributions are:
\begin{itemize}
    \item We identify two practical bottlenecks for real-world offline RL: the sensitivity induced by explicit behavior regularization and the approximation error induced by indirect latent distillation (e.g., noise aliasing).
    \item We propose \textbf{Latent Policy Steering (LPS)}, which structurally decouples behavioral constraints from reward maximization by enabling \textbf{direct} latent policy improvement via backpropagation through a differentiable one-step generative model.
    \item We demonstrate that LPS achieves state-of-the-art performance on OGBench and exhibits superior practicality on real-world robotic manipulation, consistently outperforming behavioral cloning without task-specific tuning.
\end{itemize}

\section{Related Work}
\label{sec:related_work}

\subsection{Generative Behavior Constraints in Offline RL}
\label{sec:related_regularized_rl}

TD3+BC~\citep{DBLP:conf/nips/FujimotoG21} is a widely used baseline for offline RL, but standard parametric actor can struggle to model multimodal action distributions common in robotics data. Because of this, recent methods incorporate expressive generative behavior models, including diffusion policies~\citep{DBLP:conf/iclr/WangHZ23, DBLP:conf/rss/ChiFDXCBS23} and flow-based models~\citep{DBLP:conf/icml/ParkLL25}, often combined with \emph{action chunking}~\citep{DBLP:conf/rss/ZhaoKLF23, DBLP:conf/rss/ChiFDXCBS23,li2025qc} to capture long-horizon structure. However, many of these methods rely on an explicit trade-off between policy improvement and behavior regularization, controlled by a sensitive hyperparameter, which can be difficult to tune reliably on real robots. 

An alternative direction is to adjust this trade-off at inference time. For example, CFGRL~\citep{frans2025diffusionguidancecontrollablepolicy} applies classifier-free guidance (CFG)~\citep{ho2022classifier} to an optimality-conditioned generative policy, enabling controllable interpolation between behavioral adherence and task performance even in large VLA models~\citep{intelligence2025pi06vlalearnsexperience}. In contrast, our approach performs \emph{direct} latent policy improvement using action-space Q-gradients propagated through a differentiable generative policy, preserving the structural constraints of the behavior model while maximizing expected policy return.

\subsection{Reinforcement Learning in Latent Action Spaces}

Latent action models address distributional shift in offline RL by restricting optimization to a learned manifold. Approaches such as PLAS~\citep{DBLP:conf/corl/ZhouBH20} and LAPO~\citep{lapo} use variational autoencoders (VAEs)~\citep{DBLP:journals/corr/KingmaW13} to construct a compact latent space capturing the dataset support, then optimize a policy over latents rather than the unconstrained action space to reduce extrapolation error. Related work in hierarchical and skill-based RL~\citep{DBLP:conf/iclr/AjayKALN21, DBLP:conf/corl/PertschLL20} similarly leverages latent variables to represent temporally extended behaviors.

This paradigm has recently been extended to expressive generative behavior models. DSRL~\citep{wagenmaker2025steering} steers a frozen diffusion policy by optimizing latent inputs with respect to a critic, effectively combining structural constraints with value-driven improvement. In the offline setting, its noise aliasing variant DSRL-NA introduces an additional distillation step to approximate a latent-space critic from an action-space critic. While effective in some settings, this extra approximation can limit purely-offline performance. Our method removes the need for latent critic distillation by backpropagating gradients from an action-space critic through a differentiable one-step generative policy to update the latent actor directly.

\subsection{One-Step Generative Models for Robot Learning}
\label{sec:related_onestep}

To reduce the cost of iterative denoising, recent work has pivoted towards accelerating sampling via distillation and rectification, including progressive distillation~\citep{DBLP:conf/iclr/SalimansH22}, consistency models~\citep{DBLP:journals/corr/abs-2407-02398}, rectified flow~\citep{DBLP:conf/iclr/LiuG023} and shortcut models~\citep{fransone}. These techniques have been actively adopted in robot learning to enable fast action sampling and fine-tuning. For instance, Flow Q-Learning (FQL)~\citep{DBLP:conf/icml/ParkLL25} distills a generative behavior model into a one-step policy for deterministic policy extraction, and other approaches uses consistency-style objectives for fast inference~\citep{DBLP:conf/aaai/Zhang0F0ZL25} and efficient online fine-tuning~\citep{chen2025conrft}. 

MeanFlow~\citep{geng2025meanflowsonestepgenerative} provides a differentiable one-step generative formulation that has recently been applied to robotics and RL. MeanFlowQL~\cite{wang2025onestepgenerativepoliciesqlearning} integrates a MeanFlow behavior policy into the TD3+BC framework, and MP1~\cite{sheng2025mp1meanflowtames} leverages MeanFlow for fast 3D manipulation. We build on this line of work by using MeanFlow as a differentiable mapping from latents to actions, enabling direct latent policy steering via gradients from an action-space critic.

\section{Preliminaries}

\subsection{Reinforcement Learning with Action Chunking}

Action chunking is often crucial in real-world robotics, offering both temporal coherence and improved handling of multi-modality. Following prior work, we adopt Q-Chunking (QC), introduced as part of QC-FQL offline RL algorithm~\citep{li2025qc}, to train action-chunked critics for our method and all baselines. 

Rather than predicting a single action $a_t$ at each timestep, an action-chunked policy produces a length-$h$ action sequence $a_{t:t+h}=(a_t, a_{t+1}, ..., a_{t+h-1})$ via $\pi_\phi(a_{t:t+h} \mid s_t)$, and the corresponding chunked critic is defined as $Q_\theta(s_t, a_{t:t+h})$. This formulation enables temporally coherent behavior generation in an open-loop manner. More importantly, unlike standard $n$-step returns that introduce off-policy bias, QC supports $h$-step bootstrapped value backups using in-dataset rewards:
\begin{equation}
\begin{split}
\mathcal{L}_{Q} = \mathbb{E}_{\mathcal{D}} \Big[ & Q_\theta(s_t, a_{t:t+h}) \\
 & - \left( r_{t:t+h} + \gamma^h Q_{\bar\theta}(s_{t+h}, a_{t+h:t+2h}) \right)^2 \Big],
\end{split}
\label{eq:q-chunking}
\end{equation}
where $r_{t:t+h} = \sum_{i=0}^{h-1}\gamma^i r_{t+i}$, $a_{t+h:t+2h} \sim \pi_\phi(\cdot \mid s_{t+h})$, and $\bar{\theta}$ is a target network. 

To mitigate distribution shift, QC-FQL constrains the learned policy to remain close to the offline behavior distribution. In QC-FQL, this is implemented via a squared 2-Wasserstein upper bound between the learned policy and a behavior policy. Concretely, the policy is parameterized as a one-step flow model $\pi_\phi(s, z)$, and optimized to maximize Q-value while staying close to a flow-based behavior policy $\pi_\beta(s,z)$:
\begin{equation}
\mathcal{L}_\mathrm{\mbox{\small QC-FQL}} = \underbrace{- \mathbb{E}[Q(s, \pi_\phi(s, z))]}_{\text{Extraction}} + \alpha \cdot \underbrace{\mathbb{E}[(\pi_\phi(s, z) - \pi_\beta(s, z))^2]}_{\text{Regularization}}.
\label{eq:tug_of_war}
\end{equation}
This objective encourages actions that are both high-value and close to consistent behaviors in the dataset.

\subsection{MeanFlow for One-step Generative Modeling}

In this work, we employ \textbf{MeanFlow}~\citep{geng2025meanflowsonestepgenerative} as our base generative policy. MeanFlow models average velocity (equivalently, displacement) along a probability path, enabling one-step sampling without an auxiliary loss and iterative denoising. Let $z_t$ denote an intermediate state on the probability path connecting the data distribution (at $t=0$) to a prior latent distribution (at $t=1$). Standard flow matching models the instantaneous velocity field $v(z_t, t)$, whereas MeanFlow models the \emph{average velocity} $u(z_t, r, t)$ between two time steps $r<t$:
\begin{equation}
    u(z_t, r, t) \triangleq \frac{1}{t - r} \int_{r}^{t} v(z_{\tau}, \tau) d\tau.
    \label{eq:meanflow_def}
\end{equation}
Differentiating \Cref{eq:meanflow_def} with respect to $t$ yields the \emph{MeanFlow Identity}, which relates the learnable average velocity to the instantaneous velocity:
\begin{equation}
    \underbrace{u(z_t, r, t)}_{\text{average vel.}} = \underbrace{v(z_t, t)}_{\text{instant. vel.}} - (t - r) \underbrace{\frac{d}{dt}u(z_t, r, t)}_{\text{time derivative}}.
    \label{eq:meanflow_identity}
\end{equation}
MeanFlow trains a parameterized network $u_\beta$ to satisfy \Cref{eq:meanflow_identity} by regressing to a target constructed from the right-hand side:
\begin{equation}
    \mathcal{L}_\mathrm{MF} = \mathbb{E}_{t, z_t} \left[ \lVert u_\beta(z_t, r, t) - \mathrm{sg}(u_{\mathrm{tgt}}) \rVert_2^2 \right], \label{eq:meanflow}
\end{equation}
where $\mathrm{sg}(\cdot)$ denotes stop-gradient and $u_{\mathrm{tgt}} = v(z_t, t) - (t - r) \left( v(z_t, t) \partial_z u_\beta + \partial_t u_\beta \right)$. After training, one-step sampling maps a latent $z$ (at $t=1$) to a data sample (at $t=0$), an action chunk $\hat{a}$ in our case, via a simple \textbf{one-step ODE}:
\begin{equation}
    \hat{a} = z - u_\beta(z, 0, 1).
    \label{eq:one_step_sampling}
\end{equation}
This provides a differentiable one-step generative policy that we later exploit for end-to-end gradient propagation.


\subsection{Limitations of Prior Work}

\begin{figure}[t]
    \centering
    \includegraphics[width=\linewidth]{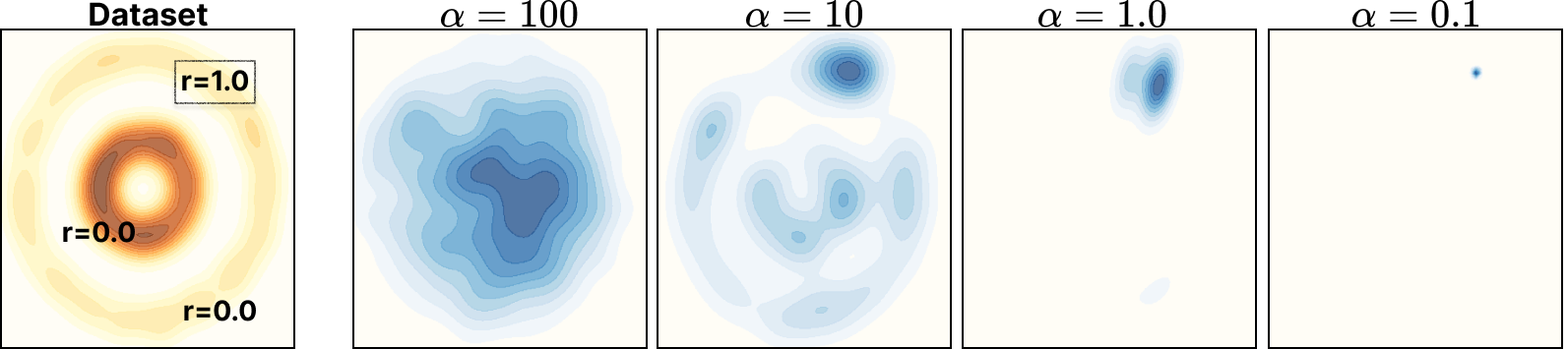}
    \caption{\textbf{Sensitivity to the regularization weight $\alpha$ in FQL.} Learned policy densities on a 2D toy task with reward concentrated in the top-right corner reveal a pattern: large $\alpha$ yields overly conservative policies, while small $\alpha$ encourages out-of-support actions.} 
    \label{fig:pre_alpha}
\end{figure}

\noindent
\textbf{Behavior regularization is sensitive to $\alpha$.} 
The way behavior-regularized offline RL methods balance value maximization and behavioral adherence -- using a weighting hyperparameter $\alpha$ in \Cref{eq:tug_of_war} -- can be fragile even in simple settings.  \Cref{fig:pre_alpha} provides an example: large $\alpha$ yields overly conservative policies, while small $\alpha$ encourages out-of-support actions. In general, appropriate $\alpha$ can vary substantially with reward scale and task characteristics. As a result, methods that rely on it often require task-specific hyperparameter sweeps, which is impractical in real-world deployments. \\

\begin{figure}[b]
    \centering
    \includegraphics[width=\linewidth]{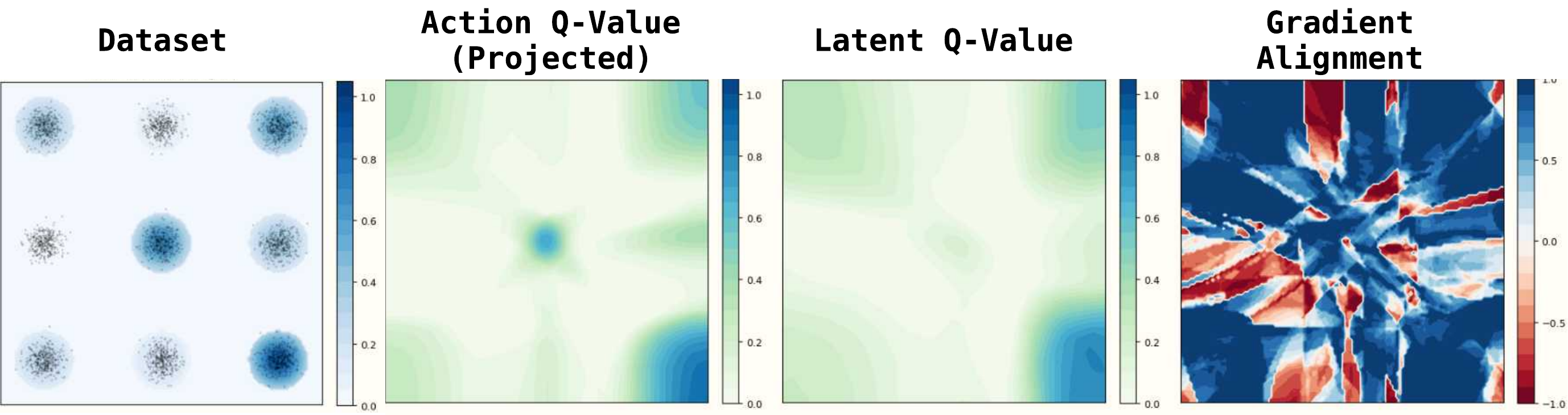}
    \caption{\textbf{Comparing action space Q-value and distilled latent-space Q-value.} 
    Left to right: (1) dataset distribution with reward intensity; 
    (2) action-space Q-value $Q_\phi(s, a)$ projected into the latent space; 
    (3) learned latent Q-value $Q_\phi(s, z)$; 
    (4) cosine similarity between the gradients in (2) and (3).}
    \label{fig:pre_latent}
\end{figure}

\noindent
\textbf{Distilled latent critics can provide poor gradients.} 
Latent steering method (e.g., DSRL) optimize latents by relying on a value function defined in the latent space. In the offline setting, this is typically obtained by distilling the action-space critic through the frozen decoder, i.e., $\min_\phi \mathbb{E} \left[ \lvert Q_\phi(s, z) - Q_\theta(\pi_\beta(s, z)) \rvert^2 \right]$. However, matching values does not guarantee that the latent gradients used for improvement are accurate. As illustrated in \Cref{fig:pre_latent}, even when $Q_\phi$ approximates values reasonably, its gradient direction $\nabla_z Q_\phi(s, z)$ can deviate substantially from the gradient of the action-space critic $\nabla_z Q_\theta(s, \pi_\beta(z))$, particularly near sharp boundaries of the data manifold. Such gradient mismatch can lead to suboptimal latent updates and degrade purely-offline performance. 

\section{Latent Policy Steering (LPS)}
\label{sec:method}


We propose \textbf{Latent Policy Steering (LPS)}, which addresses both of the above limitations. First, LPS avoids explicit behavior-regularization trade-off by \emph{separating} reward maximization and distributional constraints: a fixed generative behavior policy defines the support, while a latent actor performs value-driven steering (\textbf{resolving $\alpha$-sensitivity}). Second, LPS eliminates proxy latent critics by \emph{directly} backpropagating action-space critic gradients through a differentiable generative base policy to update the latent actor (\textbf{avoiding the inaccurate latent critic}). We instantiate LPS using three key components: a differentiable one-step base policy (\Cref{sec:base_policy}), a spherical latent geometry (\Cref{sec:latent_geometry}), and a direct latent optimization objective (\Cref{sec:optimization}).

\subsection{Differentiable Base Policy via MeanFlow}
\label{sec:base_policy}

The first component is the base policy $\pi_\beta: \mathcal{Z} \times \mathcal{S} \to \mathcal{A}$, which defines the ``safe manifold'' or the support of the dataset. While DSRL treats the base policy as a black box, LPS treats it as a \textit{differentiable mapping}. This allows us to backpropagate gradients from the action-space critic to the latent actor through $\pi_\beta$  directly. 

However, a practical obstacle is that standard diffusion or flow-matching policies typically require iterative sampling, making end-to-end backpropagation expensive and unstable. We therefore employ \textbf{MeanFlow} for the base policy, which enables efficient one-step deterministic generation. 

\textbf{Noise-to-action reformulation.}
In the original MeanFlow formulation, samples are produced by applying a learned displacement to latent noise. Early in training, errors in the displacement filed can amplify output variance, which in turn destabilizes the critic gradients used for steering. Following recent practice~\citep{li2026basicsletdenoisinggenerative,wang2025onestepgenerativepoliciesqlearning}, we use a \textbf{noise-to-action reformulation} in which $\pi_\beta$ directly predicts the denoised action (or action chunk) rather than the displacement. Concretely, we write the implied mean velocity $u_\beta$ and its time derivative as residual quantities:
\begin{equation}
    u_\beta(z_t, r, t) = z_t - \pi_\beta(z_t, r, t), \quad \frac{\mathrm{d} u_\beta}{\mathrm{d} t} = v - \frac{\mathrm{d} \pi_\beta}{\mathrm{d} t}.
    \label{eq:reform_meanflow}
\end{equation}
Substituting \Cref{eq:reform_meanflow} into the MeanFlow training objective~\Cref{eq:meanflow} yields a numerically more stable training procedure by grounding the training in the action space.

\subsection{Spherical Latent Geometry}
\label{sec:latent_geometry}

Given the base policy (mapping) $\pi_\beta$, we next define the latent space $\mathcal{Z}_\mathrm{sphere}$ where the latent actor operates. A known failure mode with unconstrained Gaussian latents is the \emph{``norm explosion''} problem. Because the latent actor is optimized to increase value without explicit bounds, it may increase $\lvert z \rvert$ to query latents that are atypical under the base policy prior, leading to out-of-distribution decoding and unstable learning.

To address this, we leverage the \emph{concentration of measure} property of high-dimensional Gaussians: for $\epsilon\sim\mathcal{N}(\mathbf{0},\mathbf{I}_d)$, most probability mass concentrates on a thin shell of radius $\sqrt{d}$~\cite{vershynin2018high}. This suggests treating the ``typical set'' of the base policy as naturally spherical. Therefore, we synchronize the support of the base policy and latent actor's output $l_\phi(s)$ by constraining both to the hypersphere $S^{d-1}$ with radius $\sqrt{d}$:
\begin{subequations}
\begin{align}
    \text{Base Policy Latent:} \quad z & \sim \sqrt{d} \cdot \frac{\epsilon}{\| \epsilon \|_2}, \quad \epsilon \sim \mathcal{N}(\mathbf{0}, \mathbf{I}_d), \label{eq:sample_sphere} \\
    \text{Latent Actor Output:} \quad & z_\phi = \pi_\phi(s) = \sqrt{d} \cdot \frac{l_\phi(s)}{\| l_\phi(s) \|_2}.\label{eq:sphere_project}
\end{align}
\label{eq:spherical_sync}
\end{subequations}
By training the base policy using latents sampled from \Cref{eq:sample_sphere} and constraining the latent actor via \Cref{eq:sphere_project}, LPS ensures that latent actor's queries always remain within the valid coverage of the base policy while maintaining well-conditioned gradients.

\begin{algorithm}[t]
\DontPrintSemicolon
\caption{Latent Policy Steering (LPS)}
\label{alg:mf_lql}

\textbf{Initialize:} base policy $\pi_\beta(s, z)$ (MeanFlow), latent actor $\pi_\phi(s)$, critic $Q_\theta(s, a)$, action chunk size $h$

\While{not converged}{
    Sample batch $\mathcal{B}=\{(s_t, a_{t:t+h}, r_{t:t+h}, s_{t+h})\} \sim \mathcal{D}$\
    
    \Comment{1. Update base policy $\pi_\beta$ (MeanFlow behavior prior)}
    
    Sample $z \sim \mathrm{Unif}(\mathcal{Z}_{\mathrm{sphere}})$ \Comment*[r]{\Cref{eq:sample_sphere}}
    
    Update $\beta$ to minimize $\mathcal{L}_\mathrm{MF}$ \Comment*[r]{\Cref{eq:meanflow},~\Cref{eq:reform_meanflow}}
    
    \BlankLine
    \Comment{2. Update latent actor $\pi_\phi$}
    
    $z_\phi \leftarrow \pi_\phi(s_t)$ \Comment*[r]{\Cref{eq:sphere_project}}
    
    $a_\mathrm{pred} \leftarrow \pi_\beta(s_t, z_\phi)$
    
    Update $\phi$ to minimize $\mathcal{L}_\mathrm{LPS}$ \Comment*[r]{\Cref{eq:lps}}
    
    \BlankLine
    \Comment{3. Update critic $Q_\theta$ (Q-Chunking)}

    Sample $z \sim \mathrm{Unif}(\mathcal{Z}_{\mathrm{sphere}})$ \Comment*[r]{\Cref{eq:sample_sphere}}

    $z_\phi' \leftarrow \pi_\phi(s_{t+h})$

    $a'_\mathrm{pred} \leftarrow \pi_\beta(s_{t+h}, z'_\phi)$

    Update $\theta$ to minimize $\mathcal{L}_{Q}$ \Comment*[r]{\Cref{eq:q-chunking}}
}
\end{algorithm}

\subsection{Direct Latent Policy Steering}
\label{sec:optimization}

Finally, we learn a latent actor $\pi_\phi: \mathcal{S}\rightarrow\mathcal{Z}$ that steers the base policy toward high-value actions. Since the base policy $\pi_\beta$ is differentiable, we can optimize $\pi_\phi$ directly using an action-space critic $Q_\theta$:
\begin{equation}
\mathcal{L}_\mathrm{LPS}= -\mathbb{E}_{s \sim \mathcal{D}} \left[ Q_\theta(s, \pi_\beta(s, \pi_\phi(s))) \right].
\label{eq:lps}
\end{equation}
Gradients of \Cref{eq:lps} propagate through $\pi_\beta$ via the chain rule, yielding low-variance latent updates without introducing proxy $Q(s, z)$.

The overall training objective of LPS sums the base-policy loss (reformulated MeanFlow), the latent steering loss, and the critic loss:
\begin{equation}
\mathcal{L}_\mathrm{Total} = \mathcal{L}_\mathrm{MF} + \mathcal{L}_\mathrm{LPS} + \mathcal{L}_{Q}.
\label{eq:overall_mflql}
\end{equation}
Notably, LPS does not require an explicit behavior-regularization coefficient  $\alpha$: behavioral constraints are enforced structurally by the fixed generative prior, whlie policy improvement is performed in the safe, synchronized latent space by maximizing the action-space critic. The full procedure is summarized in \Cref{alg:mf_lql}.

\section{Simulation Experiments}

\begin{figure}[ht]
    \centering
    \includegraphics[width=\linewidth]{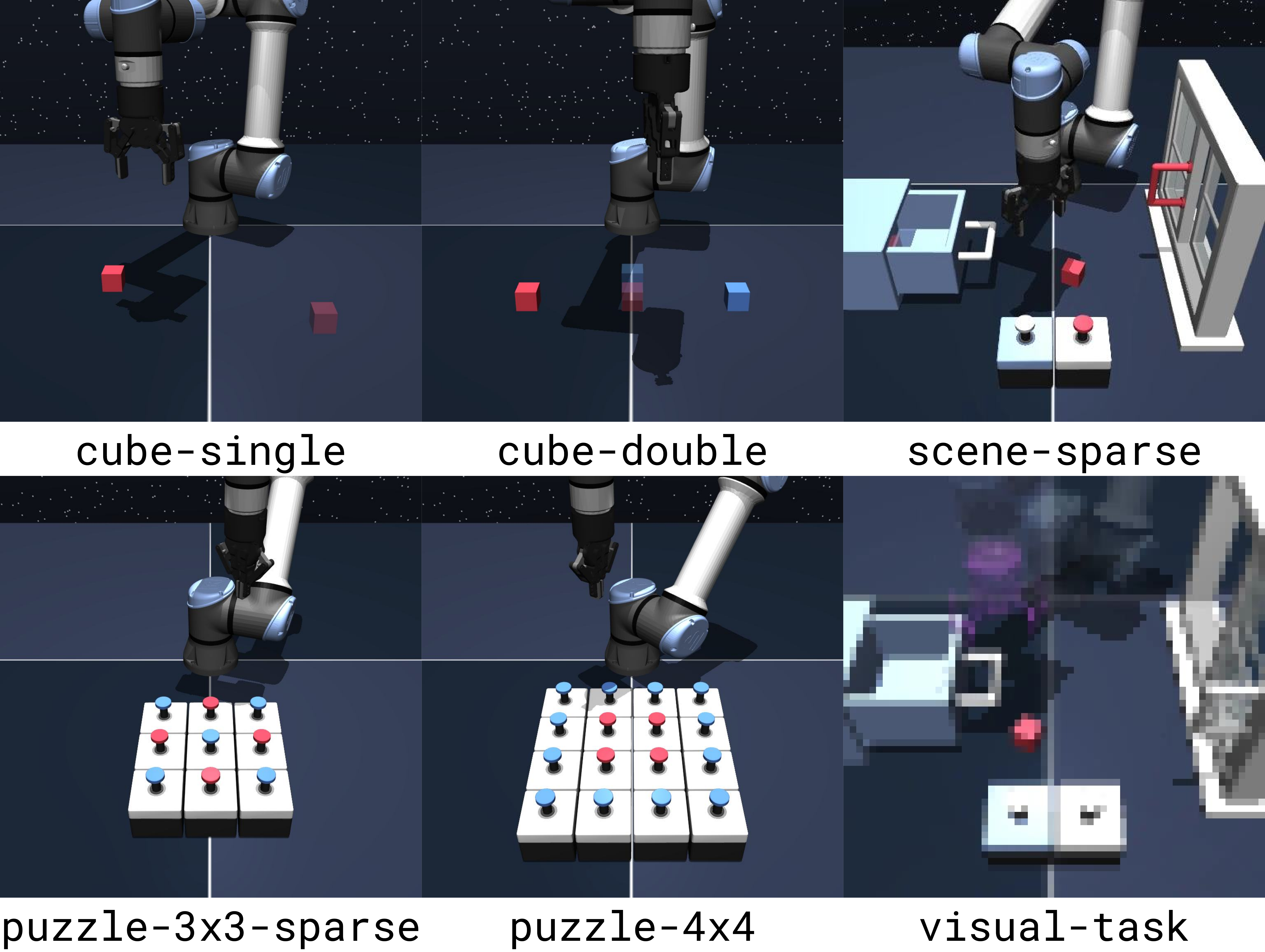}
    \caption{\textbf{OGBench Manipulation Tasks.}}
    \label{fig:ogbench_overview}
\end{figure}

\subsection{Experimental Setup}

In the simulation experiments, we evaluate  on \textbf{(1)} five \emph{state-based} manipulation tasks from OGBench~\citep{DBLP:conf/iclr/ParkFEL25}: \texttt{cube-single}, \texttt{cube-double}, \texttt{scene-sparse}, \texttt{puzzle-3x3-sparse}, and \texttt{puzzle-4x4}. Each task includes five variants corresponding to different goal configurations. We additionally consider \textbf{(2)} \emph{pixel-based} settings using the first task from each corresponding visual benchmark split~\citep{DBLP:conf/icml/ParkLL25} (denoted \texttt{visual-task}). These environments provide a rigorous testbed for isolating the effect of policy extraction under a shared value-learning algorithm. 

To focus on policy extraction mechanisms, we compare methods that all use Q-Chunking (QC)~\citep{li2025qc} for value learning,  but differ in how they represent the base policy and how they perform policy improvement:
\begin{itemize}
    \item \textbf{\textcolor{cLPS}{LPS}} (Ours): latent policy steering with a MeanFlow base policy, trained via direct backpropagation of action-space Q-gradients.

    \item \textbf{\textcolor{cQCFQL}{QC-FQL}} and \textbf{\textcolor{cQCMFQL}{QC-MFQL}}: action-space policy extraction via behavior distillation. QC-MFQL matches QC-FQL, but replaces the base policy with MeanFlow.

    \item \textbf{\textcolor{cDSRL}{DSRL}}: latent steering with a latent-space critic. For a fair comparison, we re-implement DSRL-NA using flow matching and jointly train the base policy, critic, and noise-aliasing (NA) components under the same QC value learning.

    \item \textbf{\textcolor{cCFGRL}{CFGRL}}: inference-time steering via classifier-free guidance (CFG) applied to an optimality-conditioned generative policy~\citep{frans2025diffusionguidancecontrollablepolicy}.

\end{itemize}

Across all tasks, we use chunk length $h=5$ and train each method for $1M$ gradient steps with batch size $256$. We use a $4$-layer MLP with hidden size $512$ for the base policies and $256$ for the critics. For the latent actors in \textbf{\textcolor{cLPS}{LPS}} and \textbf{\textcolor{cDSRL}{DSRL}}, we use a $2$-layer MLP with hidden size $256$. We carefully tune $\alpha$ for \textbf{\textcolor{cQCFQL}{QC-FQL}} and \textbf{\textcolor{cQCMFQL}{QC-MFQL}}. For \textbf{\textcolor{cCFGRL}{CFGRL}}, we use the best-reported CFG strength $w$ from \citep{frans2025diffusionguidancecontrollablepolicy}. Following common practice, we normalize the critic loss to have unit norm. 


\subsection{Experimental Results}

\begin{figure}[ht]
    \centering
    \includegraphics[width=0.49\textwidth]{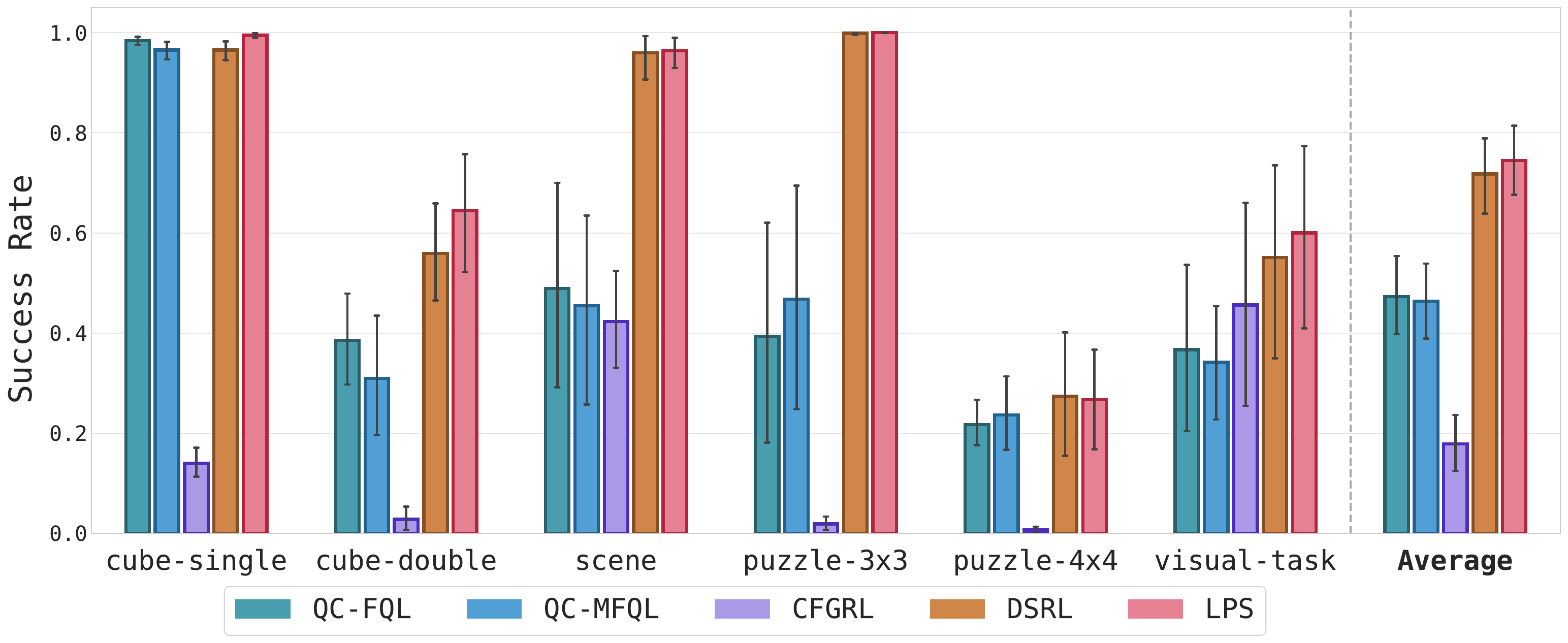}
    \caption{\textbf{Performance on OGBench.} We evaluate the success rates across tasks. Bars report the mean success rate over $3$ seeds, and error bars indicate the $95$\% confidence interval estimated using bootstrap resampling with $1$K iterations.}
    \label{fig:ogbench_learning_curve}
\end{figure}

\Cref{fig:ogbench_learning_curve} reports success rates on OGBench. Although all methods share the same QC value-learning mechanism, performance varies significantly with the policy extraction strategy. \textbf{\textcolor{cLPS}{LPS}} consistently outperforms the one-step distillation baselines (\textbf{\textcolor{cQCFQL}{QC-FQL}} and \textbf{\textcolor{cQCMFQL}{QC-MFQL}}).

\textbf{\textcolor{cDSRL}{DSRL}} exhibits higher variance across tasks and performs poorly on the challenging \texttt{cube-double} domain, highlighting the limitations of relying on a distilled latent-space critic in the offline setting. Despite being an out-of-the-box solution, \textbf{\textcolor{cCFGRL}{CFGRL}} underperforms explicit policy extraction methods, suggesting that inference-time guidance alone provides weaker and less precise improvement signals than direct critic-based optimization.

\subsection{Sensitivity to the Regularization Weight $\alpha$}

\begin{figure}[b]
    \centering
    \includegraphics[width=\linewidth]{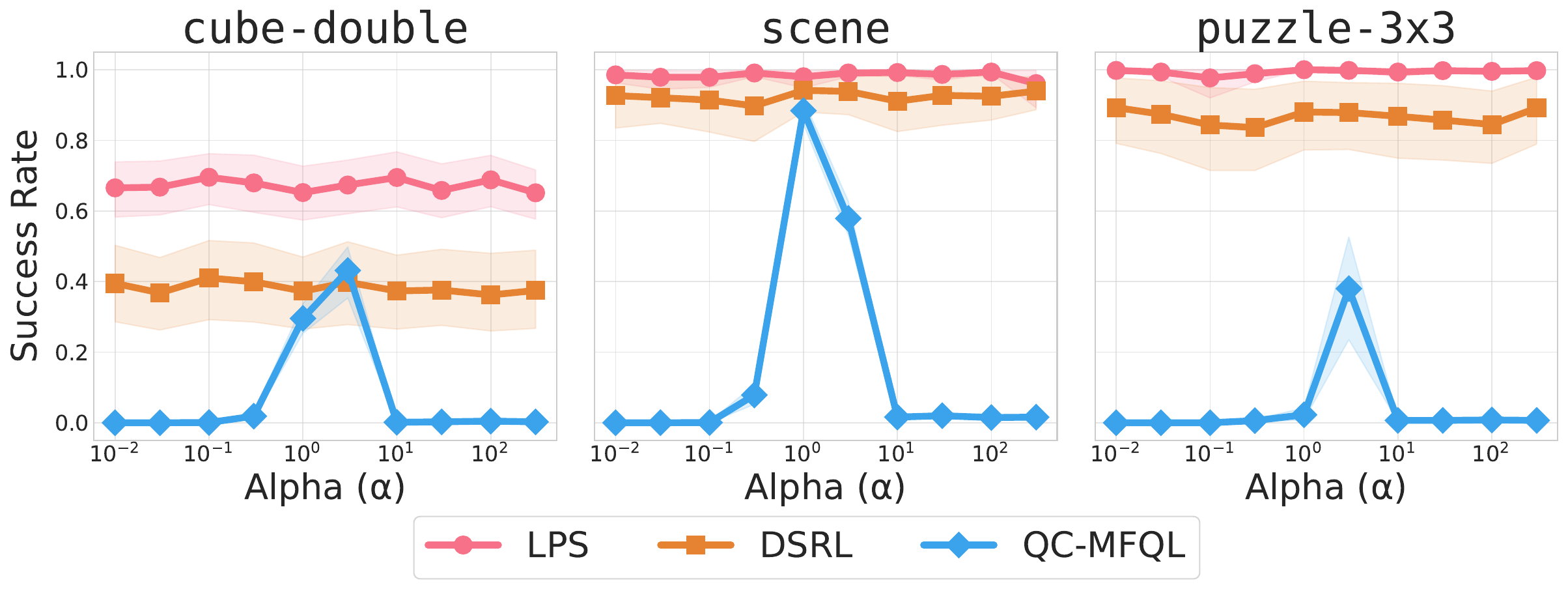}
    \caption{\textbf{Sensitivity to $\alpha$.} We report the success rates of \textbf{\textcolor{cQCMFQL}{QC-MFQL}}, \textbf{\textcolor{cDSRL}{DSRL}}, and \textbf{\textcolor{cLPS}{LPS} (Ours)} across varying $\alpha$ (swept from $0.01$ to $300$) on representative default tasks used for hyperparameter tuning. Solid lines denote the mean success rate, and shaded regions show 95\% confidence interval.}
\label{fig:alpha_sensitivity}
    \label{fig:alpha_plot}
\end{figure}

\begin{figure*}[ht]
    \centering
    \includegraphics[width=0.9\textwidth]{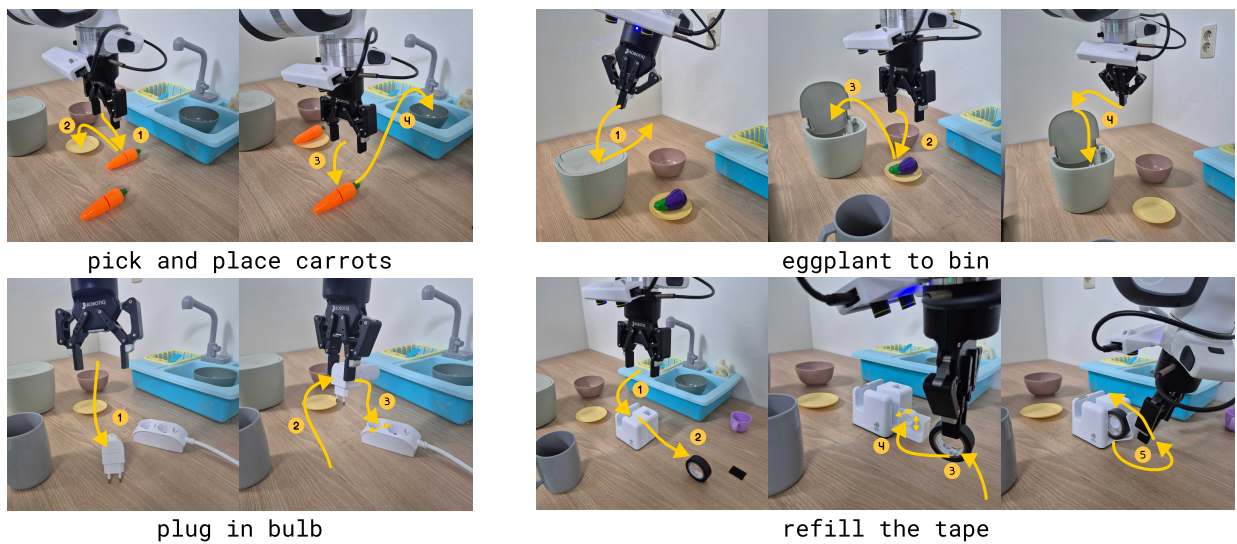}
    \caption{\textbf{Overview of real-world tasks.} 
    Our real-world benchmark spans four manipulation tasks, ranging from simple pick-and-place to precision insertion and trajectory stitching. We collect $50$ human teleoperated demonstrations per task.}
    \label{fig:real_exp}
\end{figure*}

To evaluate robustness to behavior-regularization tuning, we sweep $\alpha$ on three representative tasks used during hyperparameter tuning, as shown in \Cref{fig:alpha_sensitivity}. For both \textbf{\textcolor{cLPS}{LPS}} and \textbf{\textcolor{cDSRL}{DSRL}}, which do not inherently include $\alpha$, we weighted the base policy loss by $\alpha$ to align with the experimental setting. As expected, \textbf{\textcolor{cQCMFQL}{QC-MFQL}}, representing action-space regularization methods, exhibits a sharp performance peak at a specific $\alpha$, with success rates dropping rapidly when $\alpha$ deviates from the task-specific optimum. In stark contrast, \textbf{\textcolor{cLPS}{LPS}} remains stable across a wide range of $\alpha$, consistent with our design goal of decoupling policy improvement from explicit behavior-regularization weights. 

Meanwhile, \textbf{\textcolor{cDSRL}{DSRL}} also exhibits strong robustness to $\alpha$, supporting the intuition that latent-space optimization can be robust to behavior-regularization tuning. However, it consistently underperforms \textbf{\textcolor{cLPS}{LPS}}, suggesting that robustness to $\alpha$ alone is not sufficient. Accurate policy extraction in the offline setting benefits from directly optimizing with action-space critic gradients rather than relying on a distilled latent-space critic.

\section{Real-World Experiments}

Our goal in this section is to verify whether LPS functions as a practical, out-of-the-box solution readily applicable to real-world robotic tasks. 

\subsection{Experimental Setup}

Our simulation experiments suggest that LPS provides a robust offline RL algorithm under a shared value-learning backbone. We now evaluate whether these gains transfer to real robots. We conduct experiments on the DROID platform~\citep{DBLP:conf/rss/KhazatskyP0BDKN24} across four tasks and collect $50$ demonstrations per task (\Cref{fig:real_exp}). These tasks require high precision, closed-loop correction, and trajectory stitching--regimes where standard BC often struggles. 

We compare against \textbf{\textcolor{cDSRL}{DSRL}} as a representative latent-steering baseline in offline setting. For a fair comparison, we train its base policy, critic, and noise-aliasing  components jointly under the same training pipeline. We also include \textbf{\textcolor{cFMBC}{Flow-BC}} and \textbf{\textcolor{cMFBC}{MF-BC}}, which correspond to the underlying generative base policies trained with BC only (i.e., without RL).

We use action chunking with $h=5$ for all tasks. For the base policy, we adopt a Diffusion Transformer (DiT)~\citep{DBLP:conf/iccv/PeeblesX23} resulting $114M$ parameters and train it for $10$K gradient steps with batch size $256$. We use a semi-sparse reward, where an agent receives $-1$ per time step and $0$ upon success, with a discount factor $\gamma = 0.99$. During evaluation, an episode is terminated if the agent fails the task or exceeds the maximum horizon of $500$ steps.

\subsection{Performance Comparison}

\begin{figure}[b]
    \centering
    \includegraphics[width=\linewidth]{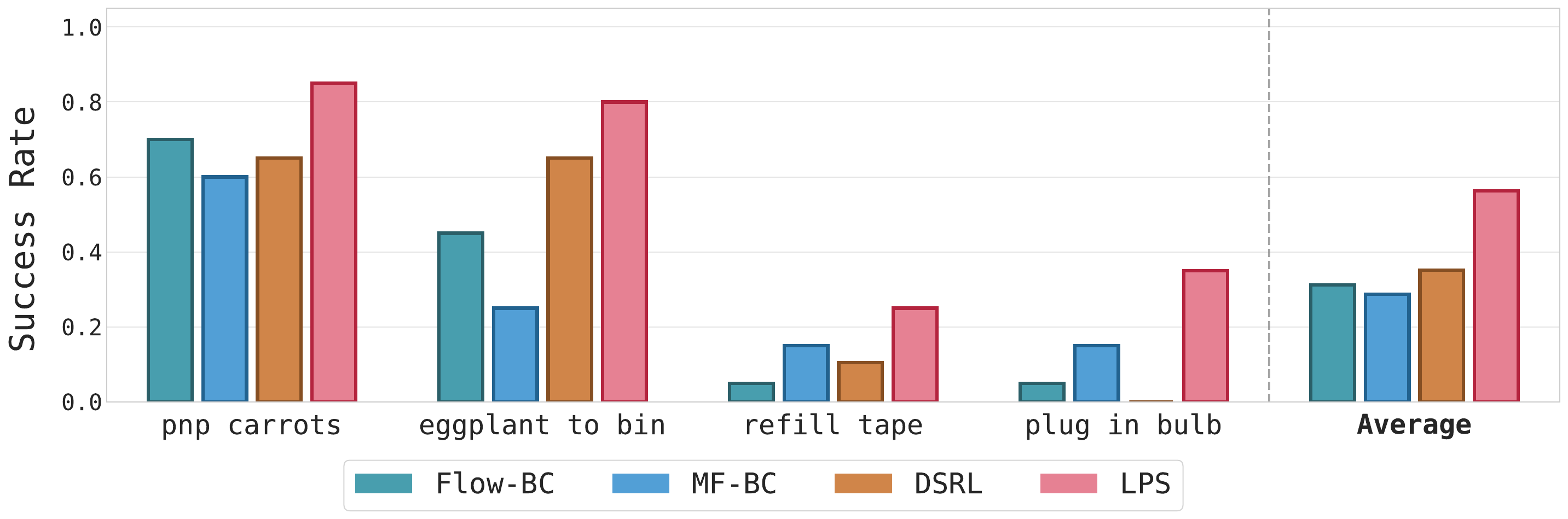}
    \caption{\textbf{Success rates on real-world tasks.} 
    We report the success rates measured over $20$ evaluation trials for each task. Our method (\textbf{\textcolor{cLPS}{LPS}}) consistently outperforms BC-based baselines and prior latent-steering methods (DSRL), demonstrating superior robustness or our method in the real world.}
    \label{fig:real_results}
\end{figure}

\Cref{fig:real_results} summarizes the real-world performance. Across all tasks, \textbf{\textcolor{cLPS}{LPS}} achieves the highest success rates and the best average performance, outperforming both behavioral cloning and prior latent-steering methods. These results indicate that directly steering a behavior policy using action-space critic gradients yields practical improvements on real robots.

\textbf{Limitations of DSRL.} 
While \textbf{\textcolor{cDSRL}{DSRL}} improves over BC on relatively simpler tasks, e.g., \texttt{pnp carrots} and \texttt{eggplant to bin}, it struggles on more challenging, precision-critical tasks. In particular, on the \texttt{plug in bulb} task, \textbf{\textcolor{cDSRL}{DSRL}} achieves 0\% success and performs worse than the base policies, suggesting that relying on a distilled latent-space critic can be fragile in purely offline deployment for challenging tasks.


\subsection{When BC Fails and How LPS Improves?}

\begin{figure}[t]
    \centering
    \includegraphics[width=\linewidth]{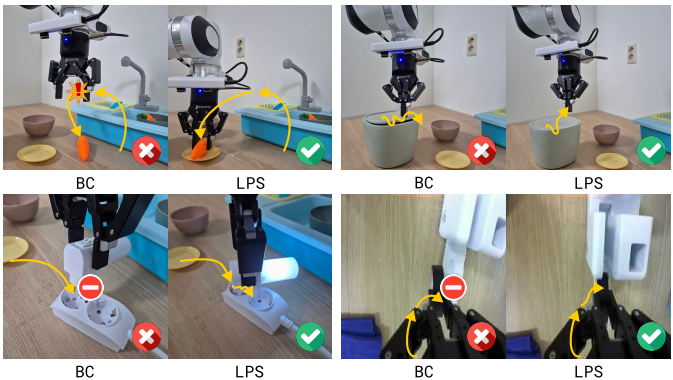}
    \caption{\textbf{Qualitative failure modes and corrections.} Teleoperation artifacts can induce failures such as premature release \textit{(top-left)}, repetitive loops \textit{(top-right)}, and freezing during alignment \textit{(bottom)}. LPS reduces these failures by selecting higher-value actions at critical decision points.}
    \label{fig:qualitative_failures}
\end{figure}

Our dataset consists solely of successful trajectories, which are often suboptimal due to human teleoperation artifacts like hesitation, micro-corrections, jittery motions, and pauses. These artifacts inherently limit the asymptotic performance of pure BC methods (\textbf{\textcolor{cFMBC}{Flow-BC}} and \textbf{\textcolor{cMFBC}{MF-BC}}). We qualitatively analyzed the resulting policy behaviors to identify specific failure modes caused by these limitations, as illustrated in \Cref{fig:qualitative_failures}. For instance, BC baselines frequently suffer from premature release due to hesitation (\texttt{pnp carrots}), repetitive motion loops (\texttt{eggplant to bin}), and freezing during precision alignment (\texttt{plug in bulb}, \texttt{refill tape}). In contrast, \textbf{\textcolor{cLPS}{LPS}} effectively mitigates these issues by steering the latent policy toward high-value regions, enabling the agent to execute decisive actions where BC baselines would otherwise stall or oscillate. While \textcolor{cLPS}{LPS} does not eliminate all failures, it substantially reduces their frequency, yielding more reliable real-world policy deployment than BC.


\begin{figure}[ht]
    \centering
    \includegraphics[width=\linewidth]{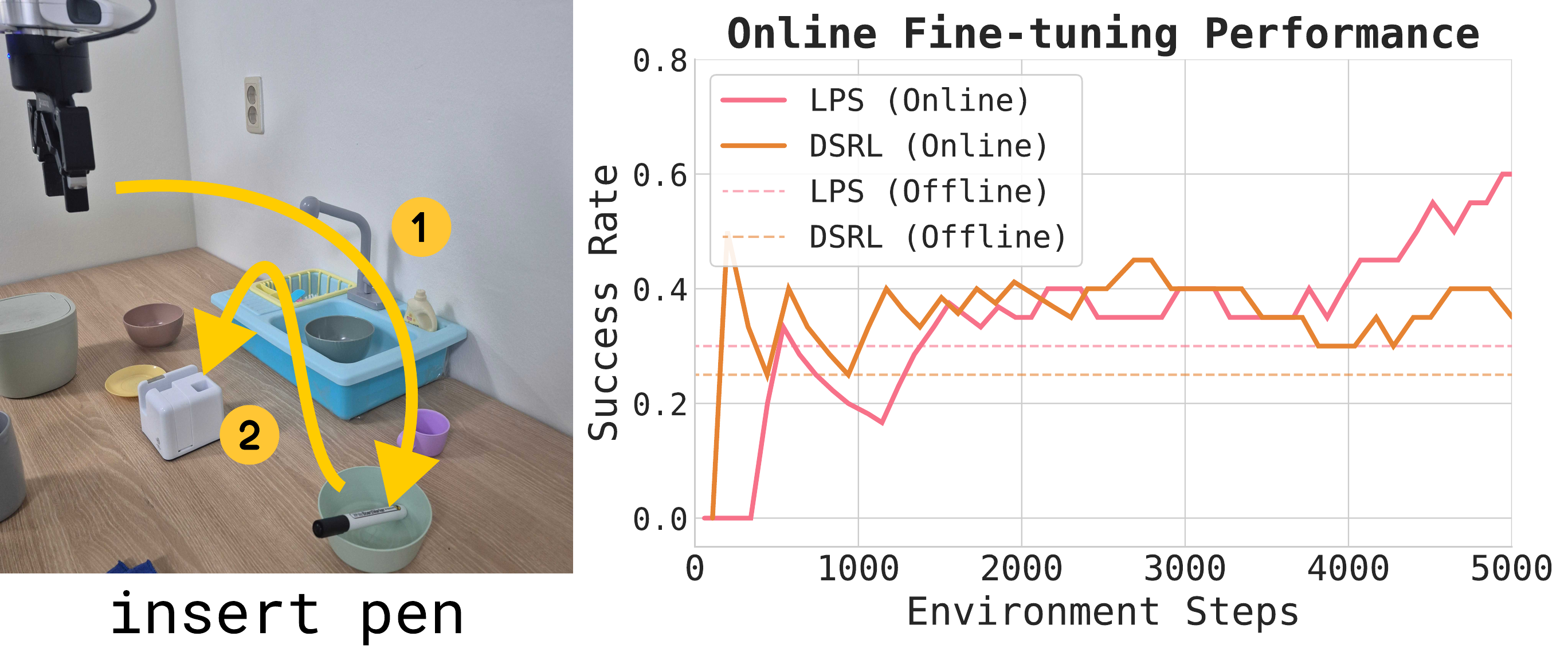}
    \caption{\textbf{Online fine-tuning results.} We evaluate online fine-tuning performance on the \texttt{insert pen} task \textit{(left)}. The learning curves \textit{(right)} show that \textbf{\textcolor{cLPS}{LPS}} efficiently improves upon its offline initialization via online interaction, surpassing \textbf{\textcolor{cDSRL}{DSRL}} within $5$K environment steps.}
    \label{fig:online}
\end{figure}

\subsection{LPS can improve via online interaction}

To demonstrate the extensibility of our framework, we investigated whether \textbf{\textcolor{cLPS}{LPS}} can be effectively applied to online fine-tuning. We conducted experiments on the \texttt{insert pen} task, initializing with offline training for $10\text{K}$ steps on a limited dataset of $20$ teleoperated demonstrations. We then fine-tuned over $5\text{K}$ environment steps. To ensure efficient learning we adopted a balanced sampling strategy: each mini-batch consisted of $64$ samples from the online replay buffer and $64$ samples from the offline dataset. We performed $200$ gradient updates between data collection rollouts, totaling $49$ and $42$ rollouts for \textbf{\textcolor{cLPS}{LPS}} and \textbf{\textcolor{cDSRL}{DSRL}}, respectively. As illustrated in \Cref{fig:online}, \textbf{\textcolor{cLPS}{LPS}} demonstrates rapid adaptation, surpassing both its offline baseline and \textbf{\textcolor{cDSRL}{DSRL}} within limited steps. This highlights the sample efficiency of our approach in leveraging online feedback.

\subsection{Computational Efficiency of LPS}

\begin{figure}[b]
    \centering
    \includegraphics[width=\linewidth]{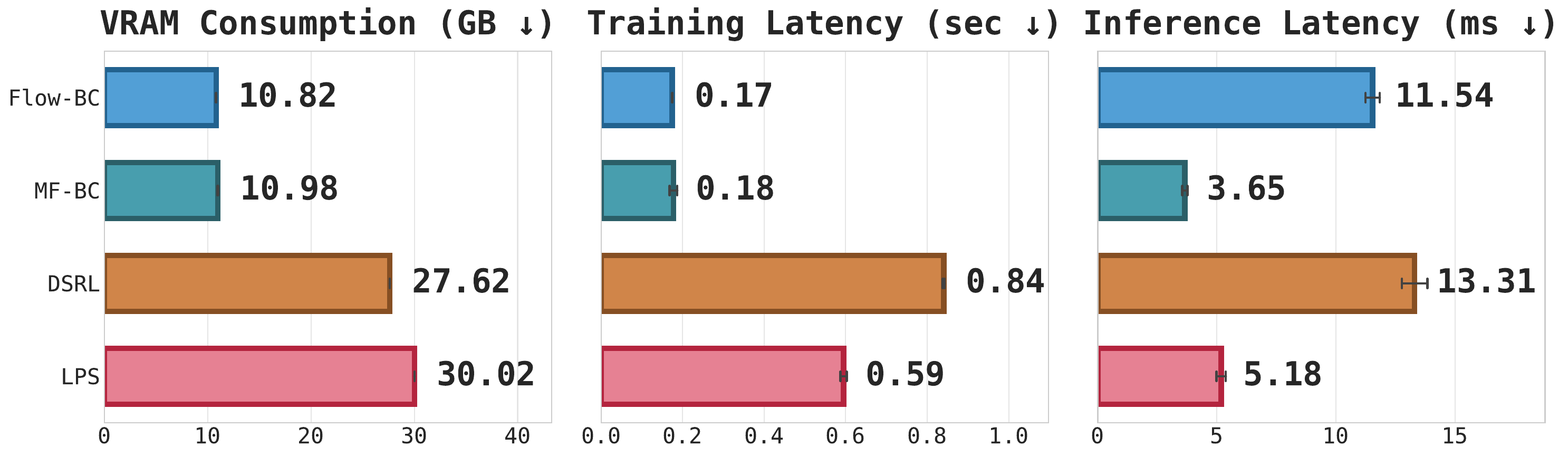}
    \caption{\textbf{Computational efficiency.} 
    We report VRAM usage and speed in training and inference. Benchmarks are measured on an NVIDIA L40S GPU with an Intel Xeon Gold 5320 CPU.}
    \label{fig:compute_eff}
\end{figure}

We also investigate computational efficiency, which is critical for real-world deployment (\Cref{fig:compute_eff}). Both \textbf{\textcolor{cDSRL}{DSRL}} and \textbf{\textcolor{cLPS}{LPS}} use more VRAM than BC baselines due to the additional critic and latent actor, yet their memory footprints are comparable to each other.
However, in terms of training speed, \textbf{\textcolor{cLPS}{LPS}} is notably faster. \textbf{\textcolor{cDSRL}{DSRL}} incurs high computational overhead from iterative sampling and noise-aliasing updates, whereas \textbf{\textcolor{cLPS}{LPS}} benefits from one-step generation and direct backpropagation through the differentiable base policy, avoiding latent-critic distillation.

At inference time, multi-step flow matching policies can introduce substantial latency. In contrast, \textbf{\textcolor{cLPS}{LPS}} utilizes MeanFlow's one-step generation, achieving inference speeds comparable to \textbf{\textcolor{cMFBC}{MF-BC}} while delivering significantly higher success rates. Overall, \textbf{\textcolor{cLPS}{LPS}} provides an attractive practical trade-off: improved performance with efficient training and fast inference.

\section{Ablation Study}

We conduct comprehensive ablation studies to validate the key design choices in \textbf{\textcolor{cLPS}{LPS}}. We focus on three components: (1) the latent-space geometry, (2) the choice of one-step generative backbone, and (3) the noise-to-action reformulation used to train MeanFlow. We follow the same evaluation protocol as in the main simulation experiments and report mean performance averaged across all state-based OGBench manipulation tasks.

\subsection{Effect of Latent-Space Geometry}
\label{sec:abl_latent}

\begin{figure*}[t]
    \centering
    \includegraphics[width=\textwidth]{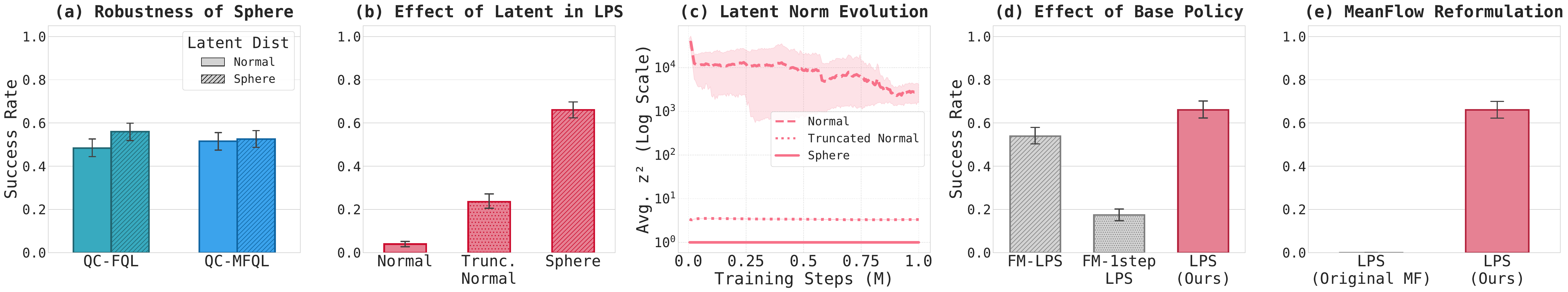}
    \caption{\textbf{Ablation studies on the key components of \textcolor{cLPS}{LPS}.} 
    \textbf{(a-c) Impact of latent-space geometry:} Success rates of (a) baseline methods and (b) \textbf{\textcolor{cLPS}{LPS}} using different latent geometries. (c) The latent norm during training, showing that the spherical constraint prevents unbounded norm growth. 
    \textbf{(d-e) Generative backbone and reformulation:} (d) Comparison of MeanFlow (\textbf{\textcolor{cLPS}{LPS}}) against flow-matching variants (10-step and 1-step sampling). (e) Performance of \textbf{\textcolor{cLPS}{LPS}} trained with the original MeanFlow objective versus the proposed noise-to-action reformulation.}
    \label{fig:ablation_latent}
\end{figure*}

We first study whether the spherical latent space retains expressivity. As shown in \Cref{fig:ablation_latent} (a), replacing the standard normal prior with a spherical prior (\texttt{sphere}) does not degrade baseline performance, suggesting that the sphere latent retains sufficient representational capacity. 

In contrast, for \textbf{\textcolor{cLPS}{LPS}}, the choice of the latent geometry is critical. We compare our method against a standard normal and a truncated normal distribution (bounded to $[-2, 2]$ and apply  $2*\tanh$ to latent actor). Both alternatives significantly reduce performance (\Cref{fig:ablation_latent}, b). The training dynamics in \Cref{fig:ablation_latent} (c) help explain why: without a spherical constraint, latent optimization tends to increase $\lvert z \rvert$ (norm growth), pushing the actor into atypical regions of the base policy. With a truncated prior, the actor often saturates near the boundary where gradients diminish. Contrasting both the base policy and actor to the same hyperspherical typical set avoids these failure modes and yields stable optimization.

\subsection{Effect of One-Step Generation Backbone}

Next, we evaluate the role of MeanFlow's one-step generation by comparing \textbf{\textcolor{cLPS}{LPS}} against two flow-matching (FM) variants: \textbf{FM-LPS}, which uses $10$ step denoising, and \textbf{FM-1step-LPS}, which forces a single Euler step at inference and during backpropagation. As shown in \Cref{fig:ablation_latent} (d), \textbf{FM-1step-LPS} performs worst, consistent with the fact that standard FM vector fields incur large approximation errors under single-step integration.
\textbf{FM-LPS} improves substantially but still underperforms our method, indicating that backpropagation through multi-step generation (i.e., BPTT through sampling trajectory) introduces additional instability and overhead that degrades learning.
Overall, these results support MeanFlow as a practical backbone for \textbf{\textcolor{cLPS}{LPS}}: it enables high-fidelity one-step generation and stable end-to-end gradients without requiring multi-step integration.

\subsection{Effect of MeanFlow Noise-to-Action Reformulation}

Finally, we ablate the noise-to-action reformulation used to train MeanFlow in our setting. \Cref{fig:ablation_latent} (e) shows that training \textbf{\textcolor{cLPS}{LPS}} with the original MeanFlow parameterization leads to unstable learning and poor downstream control, whereas the reformulated objective consistently yields strong performance. This highlights the importance of predicting denoised actions (equivalently, start-end displacement) rather than raw velocity fields.

\section{Conclusion}
\label{sec:conclusion}

In this work, we highlighted a practical bottleneck in offline RL for robotics: explicit behavior regularization can create a sensitive trade-off that often requires costly hyperparameter tuning. Latent steering offers a structural alternative, but existing offline adaptation (e.g., DSRL) typically rely on distilling an action-space critic into a latent-space critic, which can introduce approximation errors and limit purely-offline performance. To address this, we proposed \textbf{Latent Policy Steering (LPS)}, which enables direct latent policy improvement by backpropagating action-space critic gradients through a differentiable one-step base policy, together with a synchronized spherical latent geometry. Across simulation benchmarks and real-world robotic tasks, LPS consistently improves over behavioral cloning and strong latent steering baselines, providing a practical out-of-the-box approach with minimal tuning.

\textbf{Limitations.} 
LPS is ultimately bounded by the quality and coverage of the base policy. If the base policy fails to capture important modes in the data, latent steering cannot recover them. In addition, the spherical constraint is intentionally conservative. While it stabilizes optimization and keeps latent queries within  the safe, typical set of the behaviors, it may restrict extrapolation to behaviors far beyond the demonstration distribution.

\textbf{Future work.} 
Promising directions include scaling LPS to large \emph{Vision-Language-Action (VLA)} models for general-purpose robot manipulation, and exploiting the temporal structure within action chunks by using structured latent representations rather than treating chunks as flat vectors.

\section*{Acknowledgments}
This work was initiated during the first author Hokyun Im's internship at Microsoft Research Asia. This research was supported by the National Research Foundation of Korea (NRF) grant (RS-2024-00333634), the Institute of Information \& Communications Technology Planning \& Evaluation (IITP) grants (RS-2020-II201361, Artificial Intelligence Graduate School Program (Yonsei University); RS-2024-00436680, Global Research Support Program in the Digital Field Program), and the Electronics and Telecommunications Research Institute (ETRI) grant (26ZR1100, Research on Intelligent Industrial Convergence) funded by the Korean government (MSIT).


\bibliographystyle{plainnat}
\bibliography{conferences,references}

@string{ICCV = "IEEE International Conference on Computer Vision"}

@string{NeurIPS = "Neural Information Processing Systems"}

@string{ICLR = "International Conference on Learning Representations"}

@string{ICML = "International Conference on Machine Learning"}

@string{AAAI = "Association for the Advancement of Artificial Intelligence"}

@string{CoRL = "Conference on Robot Learning"}

@string{RSS = "Robotics: Science and Systems"}

@string{ICCV = "IEEE/CVF International Conference on Computer Vision"}

@string{NeurIPS = "Advances in Neural Information Processing Systems"}

@inproceedings{DBLP:conf/nips/FujimotoG21,
  author       = {Scott Fujimoto and
                  Shixiang Shane Gu},
  editor       = {Marc'Aurelio Ranzato and
                  Alina Beygelzimer and
                  Yann N. Dauphin and
                  Percy Liang and
                  Jennifer Wortman Vaughan},
  title        = {A Minimalist Approach to Offline Reinforcement Learning},
  booktitle    = NeurIPS,
  pages        = {20132--20145},
  year         = {2021},
  url          = {https://proceedings.neurips.cc/paper/2021/hash/a8166da05c5a094f7dc03724b41886e5-Abstract.html},
  bibsource    = {dblp computer science bibliography, https://dblp.org}
}

@inproceedings{DBLP:conf/iclr/WangHZ23,
  author       = {Zhendong Wang and
                  Jonathan J. Hunt and
                  Mingyuan Zhou},
  title        = {Diffusion Policies as an Expressive Policy Class for Offline Reinforcement
                  Learning},
  booktitle    = ICLR,
  publisher    = {OpenReview.net},
  year         = {2023},
  url          = {https://openreview.net/forum?id=AHvFDPi-FA},
  bibsource    = {dblp computer science bibliography, https://dblp.org}
}

@inproceedings{DBLP:conf/icml/ParkLL25,
  author       = {Seohong Park and
                  Qiyang Li and
                  Sergey Levine},
  title        = {Flow Q-Learning},
  booktitle    = ICML,
  publisher    = {OpenReview.net},
  year         = {2025},
  url          = {https://openreview.net/forum?id=KVf2SFL1pi},
  bibsource    = {dblp computer science bibliography, https://dblp.org}
}

@inproceedings{
li2025qc,
title={Reinforcement Learning with Action Chunking},
author={Qiyang Li and Zhiyuan Zhou and Sergey Levine},
booktitle=NeurIPS,
year={2025},
url={https://openreview.net/forum?id=XUks1Y96NR}
}

@inproceedings{DBLP:conf/rss/ChiFDXCBS23,
  author       = {Cheng Chi and
                  Siyuan Feng and
                  Yilun Du and
                  Zhenjia Xu and
                  Eric Cousineau and
                  Benjamin Burchfiel and
                  Shuran Song},
  editor       = {Kostas E. Bekris and
                  Kris Hauser and
                  Sylvia L. Herbert and
                  Jingjin Yu},
  title        = {Diffusion Policy: Visuomotor Policy Learning via Action Diffusion},
  booktitle    = RSS,
  year         = {2023},
  url          = {https://doi.org/10.15607/RSS.2023.XIX.026},
  doi          = {10.15607/RSS.2023.XIX.026},
  bibsource    = {dblp computer science bibliography, https://dblp.org}
}

@inproceedings{DBLP:conf/rss/ZhaoKLF23,
  author       = {Tony Z. Zhao and
                  Vikash Kumar and
                  Sergey Levine and
                  Chelsea Finn},
  editor       = {Kostas E. Bekris and
                  Kris Hauser and
                  Sylvia L. Herbert and
                  Jingjin Yu},
  title        = {Learning Fine-Grained Bimanual Manipulation with Low-Cost Hardware},
  booktitle    = RSS,
  year         = {2023},
  url          = {https://doi.org/10.15607/RSS.2023.XIX.016},
  doi          = {10.15607/RSS.2023.XIX.016},
  bibsource    = {dblp computer science bibliography, https://dblp.org}
}

@article{wagenmaker2025steering,
  author    = {Wagenmaker, Andrew and Nakamoto, Mitsuhiko and Zhang, Yunchu and Park, Seohong and Yagoub, Waleed and Nagabandi, Anusha and Gupta, Abhishek and Levine, Sergey},
  title     = {Steering Your Diffusion Policy with Latent Space Reinforcement Learning},
  journal   = CoRL,
  year      = {2025},
}

@inproceedings{DBLP:conf/iclr/SalimansH22,
  author       = {Tim Salimans and
                  Jonathan Ho},
  title        = {Progressive Distillation for Fast Sampling of Diffusion Models},
  booktitle    = ICLR,
  publisher    = {OpenReview.net},
  year         = {2022},
  url          = {https://openreview.net/forum?id=TIdIXIpzhoI},
  bibsource    = {dblp computer science bibliography, https://dblp.org}
}

@inproceedings{DBLP:conf/iclr/LiuG023,
  author       = {Xingchao Liu and
                  Chengyue Gong and
                  Qiang Liu},
  title        = {Flow Straight and Fast: Learning to Generate and Transfer Data with
                  Rectified Flow},
  booktitle    = ICLR,
  publisher    = {OpenReview.net},
  year         = {2023},
  url          = {https://openreview.net/forum?id=XVjTT1nw5z},
  bibsource    = {dblp computer science bibliography, https://dblp.org}
}

@inproceedings{DBLP:conf/aaai/Zhang0F0ZL25,
  author       = {Qinglun Zhang and
                  Zhen Liu and
                  Haoqiang Fan and
                  Guanghui Liu and
                  Bing Zeng and
                  Shuaicheng Liu},
  editor       = {Toby Walsh and
                  Julie Shah and
                  Zico Kolter},
  title        = {FlowPolicy: Enabling Fast and Robust 3D Flow-Based Policy via Consistency
                  Flow Matching for Robot Manipulation},
  booktitle    = AAAI,
  pages        = {14754--14762},
  publisher    = {{AAAI} Press},
  year         = {2025},
  url          = {https://doi.org/10.1609/aaai.v39i14.33617},
  doi          = {10.1609/AAAI.V39I14.33617},
  bibsource    = {dblp computer science bibliography, https://dblp.org}
}

@inproceedings{chen2025conrft, 
    title={Con{RFT}: A Reinforced Fine-tuning Method for {VLA} Models via Consistency Policy}, 
    author={Yuhui Chen and Shuai Tian and Shugao Liu and Yingting Zhou and Haoran Li and Dongbin Zhao}, 
    booktitle=RSS, 
    year={2025},
    doi={10.15607/RSS.2025.XXI.019}
}

@inproceedings{geng2025meanflowsonestepgenerative,
      title={Mean Flows for One-step Generative Modeling}, 
      author={Zhengyang Geng and Mingyang Deng and Xingjian Bai and J. Zico Kolter and Kaiming He},
      year={2025},
      booktitle={NeurIPS}
}

@misc{wang2025onestepgenerativepoliciesqlearning,
      title={One-Step Generative Policies with {Q}-Learning: A Reformulation of MeanFlow}, 
      author={Zeyuan Wang and Da Li and Yulin Chen and Ye Shi and Liang Bai and Tianyuan Yu and Yanwei Fu},
      year={2025},
      eprint={2511.13035},
      archivePrefix={arXiv},
      primaryClass={cs.LG},
      url={https://arxiv.org/abs/2511.13035}, 
}

@inproceedings{sheng2025mp1meanflowtames,
      title={{MP1}: Mean Flow Tames Policy Learning in 1-step for Robotic Manipulation}, 
      author={Juyi Sheng and Ziyi Wang and Peiming Li and Mengyuan Liu},
      year={2026},
      booktitle=AAAI
}

@inproceedings{DBLP:conf/rss/KhazatskyP0BDKN24,
  author       = {Alexander Khazatsky and
                  Karl Pertsch and
                  Suraj Nair and
                  Ashwin Balakrishna and
                  Sudeep Dasari and
                  Siddharth Karamcheti and
                  Soroush Nasiriany and
                  Mohan Kumar Srirama and
                  Lawrence Yunliang Chen and
                  Kirsty Ellis and
                  Peter David Fagan and
                  Joey Hejna and
                  Masha Itkina and
                  Marion Lepert and
                  Yecheng Jason Ma and
                  Patrick Tree Miller and
                  Jimmy Wu and
                  Suneel Belkhale and
                  Shivin Dass and
                  Huy Ha and
                  Arhan Jain and
                  Abraham Lee and
                  Youngwoon Lee and
                  Marius Memmel and
                  Sungjae Park and
                  Ilija Radosavovic and
                  Kaiyuan Wang and
                  Albert Zhan and
                  Kevin Black and
                  Cheng Chi and
                  Kyle Beltran Hatch and
                  Shan Lin and
                  Jingpei Lu and
                  Jean Mercat and
                  Abdul Rehman and
                  Pannag R. Sanketi and
                  Archit Sharma and
                  Cody Simpson and
                  Quan Vuong and
                  Homer Rich Walke and
                  Blake Wulfe and
                  Ted Xiao and
                  Jonathan Heewon Yang and
                  Arefeh Yavary and
                  Tony Z. Zhao and
                  Christopher Agia and
                  Rohan Baijal and
                  Mateo Guaman Castro and
                  Daphne Chen and
                  Qiuyu Chen and
                  Trinity Chung and
                  Jaimyn Drake and
                  Ethan Paul Foster and
                  Jensen Gao and
                  David Antonio Herrera and
                  Minho Heo and
                  Kyle Hsu and
                  Jiaheng Hu and
                  Donovon Jackson and
                  Charlotte Le and
                  Yunshuang Li and
                  Roy Lin and
                  Zehan Ma and
                  Abhiram Maddukuri and
                  Suvir Mirchandani and
                  Daniel Morton and
                  Tony Nguyen and
                  Abigail O'Neill and
                  Rosario Scalise and
                  Derick Seale and
                  Victor Son and
                  Stephen Tian and
                  Emi Tran and
                  Andrew E. Wang and
                  Yilin Wu and
                  Annie Xie and
                  Jingyun Yang and
                  Patrick Yin and
                  Yunchu Zhang and
                  Osbert Bastani and
                  Glen Berseth and
                  Jeannette Bohg and
                  Ken Goldberg and
                  Abhinav Gupta and
                  Abhishek Gupta and
                  Dinesh Jayaraman and
                  Joseph J. Lim and
                  Jitendra Malik and
                  Roberto Mart{\'{\i}}n{-}Mart{\'{\i}}n and
                  Subramanian Ramamoorthy and
                  Dorsa Sadigh and
                  Shuran Song and
                  Jiajun Wu and
                  Michael C. Yip and
                  Yuke Zhu and
                  Thomas Kollar and
                  Sergey Levine and
                  Chelsea Finn},
  editor       = {Dana Kulic and
                  Gentiane Venture and
                  Kostas E. Bekris and
                  Enrique Coronado},
  title        = {{DROID:} {A} Large-Scale In-The-Wild Robot Manipulation Dataset},
  booktitle    = RSS,
  year         = {2024},
  url          = {https://doi.org/10.15607/RSS.2024.XX.120},
  doi          = {10.15607/RSS.2024.XX.120},
  bibsource    = {dblp computer science bibliography, https://dblp.org}
}

@inproceedings{DBLP:conf/iccv/PeeblesX23,
  author       = {William Peebles and
                  Saining Xie},
  title        = {Scalable Diffusion Models with Transformers},
  booktitle    = {{IEEE/CVF} International Conference on Computer Vision, {ICCV} 2023,
                  Paris, France, October 1-6, 2023},
  pages        = {4172--4182},
  publisher    = {{IEEE}},
  year         = {2023},
  url          = {https://doi.org/10.1109/ICCV51070.2023.00387},
  doi          = {10.1109/ICCV51070.2023.00387},
  bibsource    = {dblp computer science bibliography, https://dblp.org}
}

@misc{li2026basicsletdenoisinggenerative,
      title={Back to Basics: Let Denoising Generative Models Denoise}, 
      author={Tianhong Li and Kaiming He},
      year={2026},
      eprint={2511.13720},
      archivePrefix={arXiv},
      primaryClass={cs.CV},
      url={https://arxiv.org/abs/2511.13720}, 
}

@inproceedings{DBLP:conf/iclr/ParkFEL25,
  author       = {Seohong Park and
                  Kevin Frans and
                  Benjamin Eysenbach and
                  Sergey Levine},
  title        = {OGBench: Benchmarking Offline Goal-Conditioned {RL}},
  booktitle    = ICLR,
  publisher    = {OpenReview.net},
  year         = {2025},
  url          = {https://openreview.net/forum?id=M992mjgKzI},
  bibsource    = {dblp computer science bibliography, https://dblp.org}
}

@inproceedings{DBLP:conf/corl/ZhouBH20,
  author       = {Wenxuan Zhou and
                  Sujay Bajracharya and
                  David Held},
  editor       = {Jens Kober and
                  Fabio Ramos and
                  Claire J. Tomlin},
  title        = {{PLAS:} Latent Action Space for Offline Reinforcement Learning},
  booktitle    = CoRL,
  series       = {Proceedings of Machine Learning Research},
  volume       = {155},
  pages        = {1719--1735},
  publisher    = {{PMLR}},
  year         = {2020},
  url          = {https://proceedings.mlr.press/v155/zhou21b.html},
  bibsource    = {dblp computer science bibliography, https://dblp.org}
}

@inproceedings{DBLP:conf/corl/PertschLL20,
  author       = {Karl Pertsch and
                  Youngwoon Lee and
                  Joseph J. Lim},
  editor       = {Jens Kober and
                  Fabio Ramos and
                  Claire J. Tomlin},
  title        = {Accelerating Reinforcement Learning with Learned Skill Priors},
  booktitle    = CoRL,
  series       = {Proceedings of Machine Learning Research},
  volume       = {155},
  pages        = {188--204},
  publisher    = {{PMLR}},
  year         = {2020},
  url          = {https://proceedings.mlr.press/v155/pertsch21a.html},
  bibsource    = {dblp computer science bibliography, https://dblp.org}
}

@book{vershynin2018high,
  title={High-dimensional probability: An introduction with applications in data science},
  author={Vershynin, Roman},
  volume={47},
  year={2018},
  publisher={Cambridge university press}
}

@misc{frans2025diffusionguidancecontrollablepolicy,
      title={Diffusion Guidance Is a Controllable Policy Improvement Operator}, 
      author={Kevin Frans and Seohong Park and Pieter Abbeel and Sergey Levine},
      year={2025},
      eprint={2505.23458},
      archivePrefix={arXiv},
      primaryClass={cs.LG},
      url={https://arxiv.org/abs/2505.23458}, 
}

@article{ho2022classifier,
  title={Classifier-free diffusion guidance},
  author={Ho, Jonathan and Salimans, Tim},
  journal={arXiv preprint arXiv:2207.12598},
  year={2022}
}

@inproceedings{fransone,
  title={One Step Diffusion via Shortcut Models},
  author={Frans, Kevin and Hafner, Danijar and Levine, Sergey and Abbeel, Pieter},
  booktitle={ICLR},
  year={2025}
}

@misc{intelligence2025pi06vlalearnsexperience,
      title={$\pi^{*}_{0.6}$: a {VLA} That Learns From Experience}, 
      author={Physical Intelligence and Ali Amin and Raichelle Aniceto and Ashwin Balakrishna and Kevin Black and Ken Conley and Grace Connors and James Darpinian and Karan Dhabalia and Jared DiCarlo and Danny Driess and Michael Equi and Adnan Esmail and Yunhao Fang and Chelsea Finn and Catherine Glossop and Thomas Godden and Ivan Goryachev and Lachy Groom and Hunter Hancock and Karol Hausman and Gashon Hussein and Brian Ichter and Szymon Jakubczak and Rowan Jen and Tim Jones and Ben Katz and Liyiming Ke and Chandra Kuchi and Marinda Lamb and Devin LeBlanc and Sergey Levine and Adrian Li-Bell and Yao Lu and Vishnu Mano and Mohith Mothukuri and Suraj Nair and Karl Pertsch and Allen Z. Ren and Charvi Sharma and Lucy Xiaoyang Shi and Laura Smith and Jost Tobias Springenberg and Kyle Stachowicz and Will Stoeckle and Alex Swerdlow and James Tanner and Marcel Torne and Quan Vuong and Anna Walling and Haohuan Wang and Blake Williams and Sukwon Yoo and Lili Yu and Ury Zhilinsky and Zhiyuan Zhou},
      year={2025},
      eprint={2511.14759},
      archivePrefix={arXiv},
      primaryClass={cs.LG},
      url={https://arxiv.org/abs/2511.14759}, 
}

@inproceedings{lapo,
 author = {Chen, Xi and Ghadirzadeh, Ali and Yu, Tianhe and Wang, Jianhao and Gao, Alex Yuan and Li, Wenzhe and Bin, Liang and Finn, Chelsea and Zhang, Chongjie},
 booktitle = NeurIPS,
 editor = {S. Koyejo and S. Mohamed and A. Agarwal and D. Belgrave and K. Cho and A. Oh},
 pages = {36902--36913},
 publisher = {Curran Associates, Inc.},
 title = {{LAPO}: Latent-Variable Advantage-Weighted Policy Optimization for Offline Reinforcement Learning},
 url = {https://proceedings.neurips.cc/paper_files/paper/2022/file/efb2072a358cefb75886a315a6fcf880-Paper-Conference.pdf},
 volume = {35},
 year = {2022}
}

@inproceedings{DBLP:journals/corr/KingmaW13,
  author       = {Diederik P. Kingma and
                  Max Welling},
  editor       = {Yoshua Bengio and
                  Yann LeCun},
  title        = {Auto-Encoding Variational Bayes},
  booktitle    = ICLR,
  year         = {2014},
  url          = {http://arxiv.org/abs/1312.6114},
  bibsource    = {dblp computer science bibliography, https://dblp.org}
}

@inproceedings{DBLP:conf/iclr/AjayKALN21,
  author       = {Anurag Ajay and
                  Aviral Kumar and
                  Pulkit Agrawal and
                  Sergey Levine and
                  Ofir Nachum},
  title        = {{OPAL:} Offline Primitive Discovery for Accelerating Offline Reinforcement
                  Learning},
  booktitle    = ICLR,
  publisher    = {OpenReview.net},
  year         = {2021},
  url          = {https://openreview.net/forum?id=V69LGwJ0lIN},
  bibsource    = {dblp computer science bibliography, https://dblp.org}
}

@inproceedings{kim2025deasdetachedvaluelearning,
      title={DEAS: DEtached value learning with Action Sequence for Scalable Offline RL}, 
      author={Changyeon Kim and Haeone Lee and Younggyo Seo and Kimin Lee and Yuke Zhu},
      year={2026},
      booktitle=ICLR
}

@article{DBLP:journals/corr/abs-2407-02398,
  author       = {Ling Yang and
                  Zixiang Zhang and
                  Zhilong Zhang and
                  Xingchao Liu and
                  Minkai Xu and
                  Wentao Zhang and
                  Chenlin Meng and
                  Stefano Ermon and
                  Bin Cui},
  title        = {Consistency Flow Matching: Defining Straight Flows with Velocity Consistency},
  journal      = {CoRR},
  volume       = {abs/2407.02398},
  year         = {2024},
  url          = {https://doi.org/10.48550/arXiv.2407.02398},
  doi          = {10.48550/ARXIV.2407.02398},
  eprinttype    = {arXiv},
  eprint       = {2407.02398},
  bibsource    = {dblp computer science bibliography, https://dblp.org}
}

\clearpage

\appendices

\section{Offline-to-Online Reinforcement Learning}

\begin{figure}[ht]
    \centering
    \includegraphics[width=\linewidth]{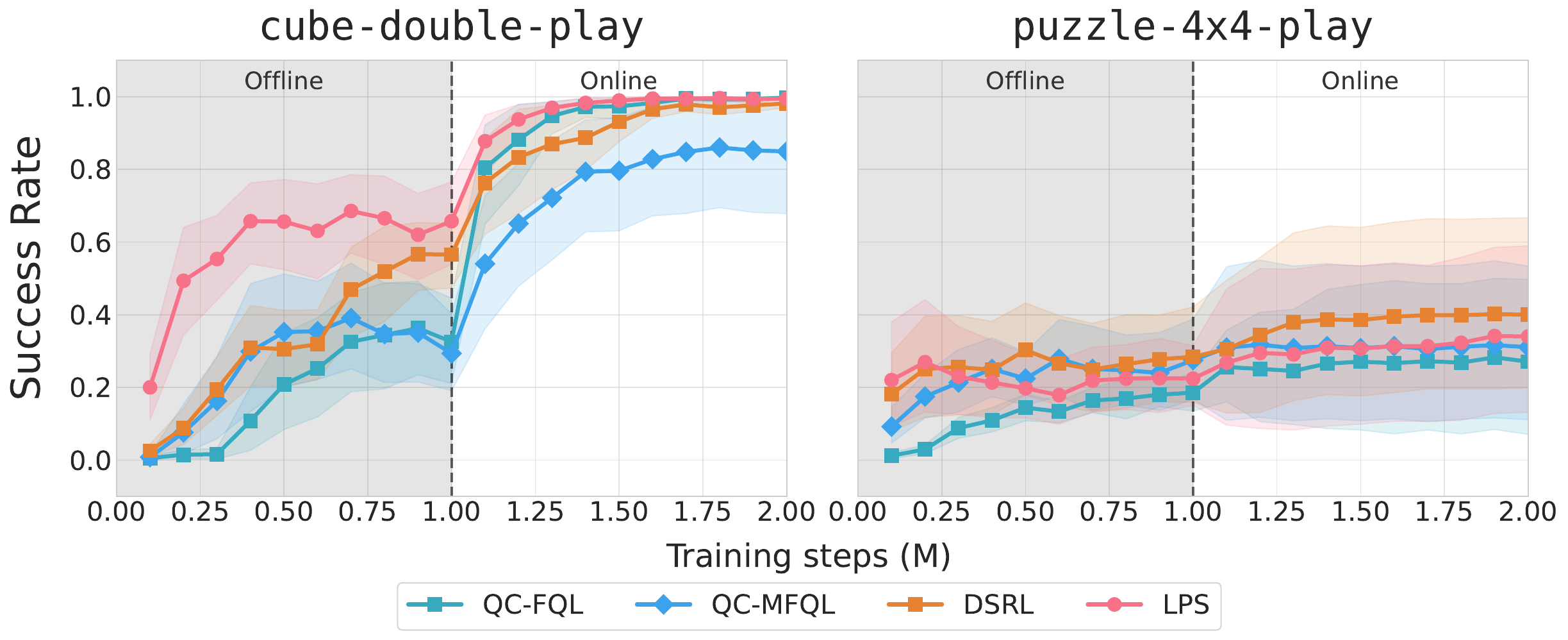}
    \caption{\textbf{Offline-to-online learning curves of LPS and baselines on OGBench tasks.}}
    \label{fig:o2o_app_learning_curves}
\end{figure}

To evaluate the adaptability and sample efficiency of LPS in a semi-offline setting, we conducted extra offline-to-online fine-tuning experiments on OGBench~\Citep{DBLP:conf/iclr/ParkFEL25}. We first pre-trained the agents using the static offline dataset for 1M gradient steps. Subsequently, the agents were deployed into the online environment to interact and collect new experiences for an additional 1M steps. This setup mimics a realistic deployment scenario where an agent is initialized with a prior policy derived from offline data and then refined via online interaction, consistent with the evaluation protocol of Q-Chunking~\Citep{li2025qc}. Figure~\ref{fig:o2o_app_learning_curves} illustrates the learning curves on \texttt{cube-double-play} and \texttt{puzzle-4x4-play}, averaged over 5 tasks and 3 seeds per domain. The vertical dashed line at 1M steps marks the transition from offline pre-training to online fine-tuning.

LPS enters the online phase from the strongest starting point and makes the most of it. On \texttt{cube-double-play}, LPS ends offline pre-training well ahead of all baselines. Upon switching to online interaction, it retains this performance without degradation and improves immediately, achieving the highest success rate throughout the early online phase. Most methods eventually saturate near perfect performance by 2M steps, so the advantage of LPS lies in reaching this level with fewer online interactions.

On \texttt{puzzle-4x4-play}, all methods improve only moderately during the online phase and end within overlapping confidence intervals. In this domain, outcomes are largely determined per run: every method solves some tasks almost completely through online interaction and remains near zero on others, so a single run can shift the domain average substantially. Indeed, the gap between DSRL~\Citep{wagenmaker2025steering} and LPS stems almost entirely from a single task--seed pair in which DSRL succeeds and LPS does not; excluding this pair, the two methods perform on par. Although the primary focus of this work is to establish a robust framework for immediate offline deployment, these findings indicate that LPS also serves as a reliable initialization for subsequent fine-tuning, providing the strongest offline starting point and the fastest online improvement on \texttt{cube-double-play} while remaining competitive on \texttt{puzzle-4x4-play}.

\section{Detailed Experimental Results}

\begin{figure}[ht]
    \centering
    \includegraphics[width=\linewidth]{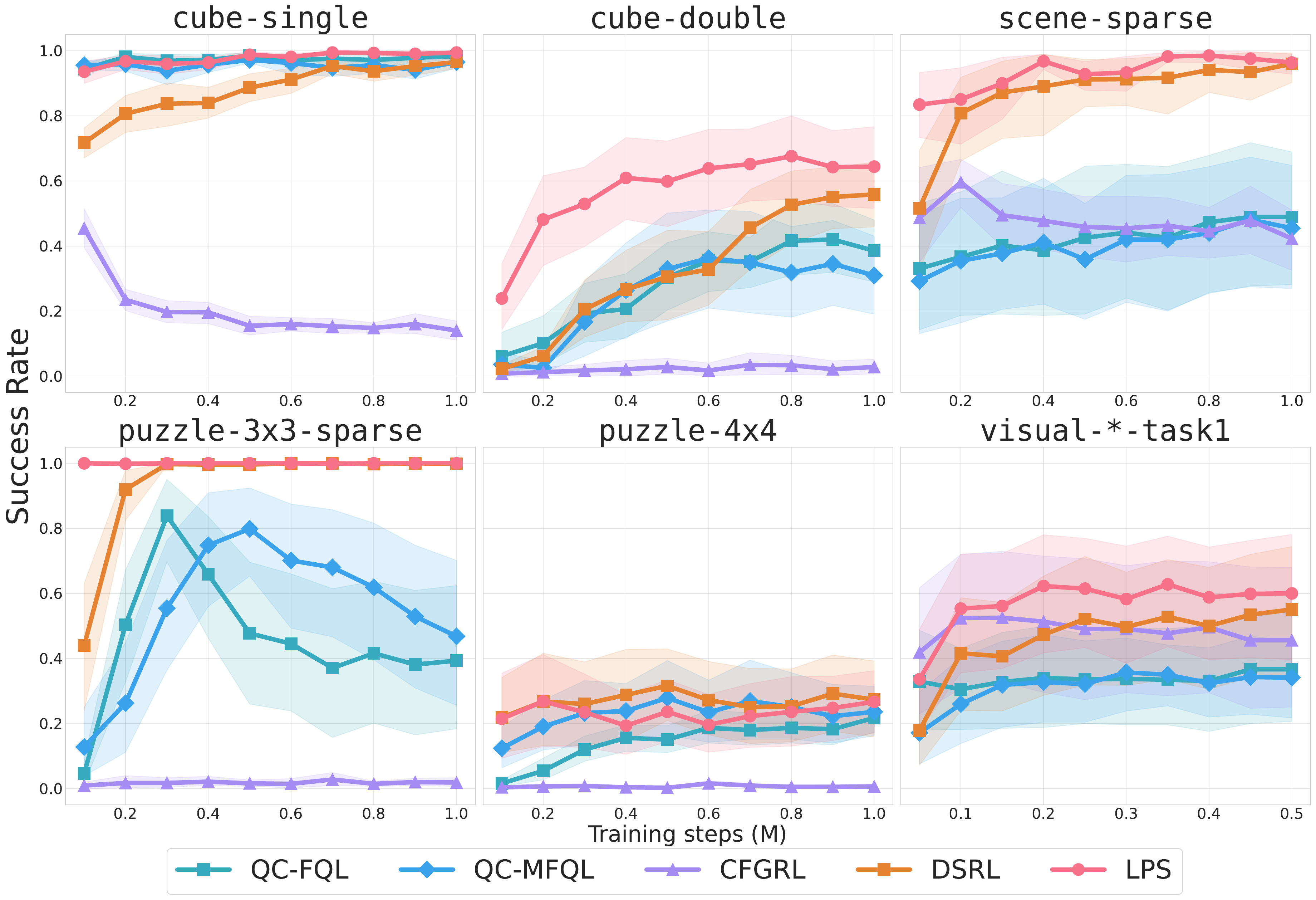}
    \caption{\textbf{Learning curves of LPS and baselines on OGBench tasks.}}
    \label{fig:app_learning_curves}
\end{figure}

\textbf{OGBench.} We report the learning curves of reported results at \Cref{fig:ogbench_learning_curve} averaged on 3 differet seeds described in \Cref{fig:app_learning_curves}. The detailed numerical results are presented in \Cref{tab:ogbench_results}. We applied two different latent space distribution (\texttt{Normal}, \texttt{Sphere}) for QC-FQL and QC-MFQL for ablation. Specifically, QC-FQL and QC-MFQL suffer from unstable training trajectories and fail to match the performance of latent actor-based methods like DSRL and LPS. Although DSRL shows strong results, it remains inferior to LPS, reaffirming the efficacy of the policy extraction mechanism employed in LPS. We also included BC baselines for comparison, and as evidenced by the results, they failed completely.

\begin{table}[b]
\centering
\caption{\textbf{Success rates on real-world tasks.}}
\label{tab:real_results}
\resizebox{\linewidth}{!}{%
\begin{tabular}{lcccc}
\toprule
Task & \textsc{BC-FM} & \textsc{BC-MF} & \textsc{DSRL} & \textbf{\textsc{LPS}} \\
    \midrule
        \texttt{eggplant to bin} & 9/20 (45\%) & 5/20 (25\%) & 13/20 (65\%) & \textbf{16/20 (80\%)} \\
        \texttt{pnp carrots} & 14/20 (70\%) & 12/20 (60\%) & 13/20 (65\%) & \textbf{17/20 (85\%)} \\
        \texttt{plug in bulb} & 1/20 (5\%) & 3/20 (15\%) & 0/20 (0\%) & \textbf{7/20 (35\%)} \\
        \texttt{refill tape} & 1/20 (5\%) & 3/20 (15\%) & 2/20 (10\%) & \textbf{5/20 (25\%)} \\
    \midrule
        \textbf{Average} & 25/80 (31.2\%) & 23/80 (28.7\%) & 28/80 (35.0\%) & \textbf{45/80 (56.2\%)} \\
    \bottomrule
\end{tabular}%
}
\end{table}

\textbf{Real world.} We report the success rates over 20 rollouts per task, appeared in \Cref{tab:real_results}. While LPS demonstrates superior performance over BC as an offline RL algorithm, it is not entirely free from limitations. It occasionally exhibits failure modes similar to those of BC, indicating room for improvement in areas such as policy and value learning.

\begin{table*}[t]
\centering
\caption{\textbf{Success rates on OGBench tasks.} Each cell represents mean ± SEM for both normal and sphere latent distributions across 3 different seeds.}
\label{tab:ogbench_results}
\resizebox{\textwidth}{!}{%
\begin{tabular}{lcccccccc}
\toprule
Task (\texttt{normal} / \texttt{sphere}) & \textsc{BC-FM} & \textsc{BC-MF} & \textsc{QC-FQL} & \textsc{QC-MFQL} &  \textsc{CFGRL} & \textsc{DSRL} & \textbf{\textsc{LPS}} \\
\midrule
\texttt{cube-single-play-singletask} & $9 \pm 1$ / $9 \pm 1$ & $10 \pm 2$ / $9 \pm 1$ & $98 \pm 0$ / $97 \pm 1$ & $97 \pm 1$ / $93 \pm 2$ & $14 \pm 2$ / - & $0 \pm 0$ / $97 \pm 1$ & \textbf{$0 \pm 0$ / $99 \pm 0$} \\
\texttt{cube-double-play-singletask} & $1 \pm 1$ / $1 \pm 1$ & $2 \pm 1$ / $2 \pm 1$ & $39 \pm 5$ / $36 \pm 6$ & $31 \pm 6$ / $40 \pm 8$ & $3 \pm 1$ / - & $0 \pm 0$ / $56 \pm 5$ & \textbf{$0 \pm 0$ / $64 \pm 7$} \\
\texttt{scene-play-sparse-singletask} & $5 \pm 2$ / $4 \pm 1$ & $3 \pm 1$ / $3 \pm 1$ & $49 \pm 11$ / $87 \pm 6$ & $45 \pm 10$ / $42 \pm 11$ & $42 \pm 5$ / - & $0 \pm 0$ / $96 \pm 3$ & \textbf{$0 \pm 0$ / $96 \pm 2$} \\
\texttt{puzzle-3x3-play-sparse-singletask} & $2 \pm 1$ / $1 \pm 1$ & $1 \pm 1$ / $1 \pm 1$ & $39 \pm 12$ / $39 \pm 12$ & $47 \pm 12$ / $42 \pm 10$ & $2 \pm 1$ / - & $0 \pm 0$ / $100 \pm 0$ & \textbf{$14 \pm 7$ / $100 \pm 0$} \\
\texttt{puzzle-4x4-play-singletask} & $0 \pm 0$ / $0 \pm 0$ & $0 \pm 0$ / $0 \pm 0$ & $22 \pm 2$ / $22 \pm 5$ & \textbf{$24 \pm 4$ / $30 \pm 6$} & $1 \pm 0$ / - & $0 \pm 0$ / $27 \pm 6$ & $3 \pm 1$ / $27 \pm 5$ \\
\texttt{visual-*-task1} & - & - & $37 \pm 9$ / - & $34 \pm 6$ / - & $46 \pm 11$ / - & - / $55 \pm 10$ & \textbf{- / $60 \pm 10$} \\
\bottomrule
\end{tabular}%
}
\end{table*}

\section{Experimental details}
\label{sec:experimental_details}

\subsection{Detailed explanation on each domain}

\textbf{OGBench.} We adapt OGBench for standard reward-maximizing offline RL by employing its single-task, reward-based variants. Specifically, we focus on the manipulation domain, which comprises five environments: \texttt{cube-single}, \texttt{cube-double}, \texttt{scene}, \texttt{puzzle-3x3}, and \texttt{puzzle-4x4}. Each environment consists of five distinct single-task configurations, yielding a total of 25 state-based tasks. Additionally, we evaluate five visual manipulation tasks ($64 \times 64 \times 3$ pixel observations) corresponding to the first configuration of each environment. The default reward is defined as $-n$ per step and $0$ upon success, where $n$ represents the number of remaining sub-goals (e.g., unlit bulbs in \texttt{puzzle} or unmatched cubes in \texttt{cube}). However, for \texttt{scene} and \texttt{puzzle-3x3}, we adopt a \textit{sparse} reward structure ($-1$ per step, $0$ upon success). We empirically found that the sub-goal-based dense rewards in these environments were not consistently aligned with the final objective, whereas the sparse setting led to superior performance.

\begin{figure}[ht]
    \centering
    \includegraphics[width=0.9\linewidth]{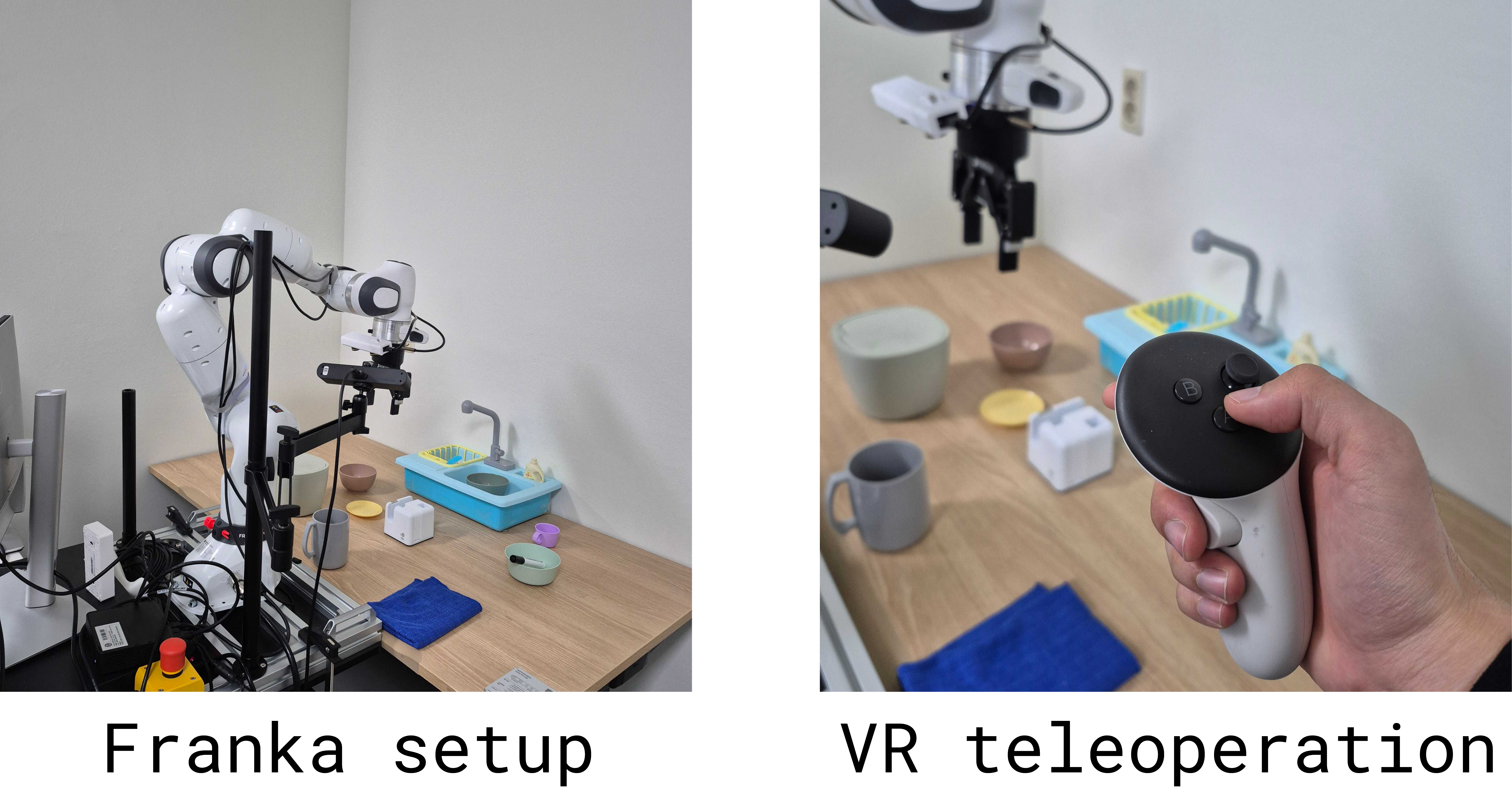}
    \caption{\textbf{Real-world setup overview.}}
    \label{fig:app_droid}
\end{figure}

\textbf{Real-world Franka.} In our real-world experiments, we strictly adhered to the DROID~\Citep{DBLP:conf/rss/KhazatskyP0BDKN24} hardware configuration. The setup comprises a Franka Research 3 robot and two ZED2 cameras, providing both third-person and wrist views ($224 \times 224 \times 3$), as shown in \cref{fig:app_droid} (Left). We employed delta end-effector control combined with a binary gripper, resulting in a 7-dimensional action space (6 dimensions for end-effector velocity and 1 for gripper actuation). For proprioceptive information, we utilize the end-effector pose and gripper state. We collected 50 human-teleoperated demonstrations using a Meta Quest 3 headset, following the DROID data collection system \cref{fig:app_droid} (Right). The collected data is stored in HDF5 format. We load all the files to the memory and concatenate it, allowing us to utilize a data loading pipeline consistent with OGBench.

\subsection{Evaluation protocol}

\textbf{OGBench.} We run 3 seeds on each OGBench task. All plots use 95\% confidence interval with stratified sampling (1000 samples). The success rate is computed by running the policy in the environment for 50 episodes and record the number of times that the policy succeeds at solving the task (and divide it by 50).

\textbf{Real-world Franka.} For all methods, we run 20 episodes on each task and calculated the success rate. We terminate each rollout when the robot stopped moving, or reach $500$ environment step.

\section{Implementation details}
\label{sec:implementation_details}

\subsection{Computational resources}
We use NVIDIA RTX-3090 GPU to run all our OGBench experiments, L40S for training real world policies, and RTX-5090 for inference and online fine-tuning.

\subsection{Baseline Implementation}
LPS and all baseline methods are built upon the \textbf{Q-chunking} codebase. We adopted the dataloading pipeline from DEAS~\citep{kim2025deasdetachedvaluelearning}. This version is simpler because it filters and samples valid data directly within the dataloader, eliminating the need for a validity mask.

\textbf{QC-MFQL.} We adapted the QC-FQL framework by modifying the base policy learning component to utilize a Mean Flow objective and employing one-step ODE sampling. We use a \textit{normal} prior distribution for training both the base policy and the one-step distillation policy.

\textbf{DSRL.} We employ the DSRL-NA variant, replacing the one-step policy distillation of QC-FQL with latent-space critic distillation and latent actor optimization. While the original DSRL incorporates an entropy term for exploration, we omit this term to align with the standard actor-critic formulation used in other offline RL baselines. Regarding the latent space structure, we replace the original $\tanh$-bounded actor with a spherical distribution (\textit{sphere}). We adopted this spherical structure as we found it to be more efficient, a finding consistent with our observations for LPS.

\textbf{CFGRL~\citep{frans2025diffusionguidancecontrollablepolicy}.} We strictly follow the original implementation of CFGRL and use the classifier-free guidance strength $w$ reported in the original paper.

For the DiT~\Citep{DBLP:conf/iccv/PeeblesX23} architecture, we leveraged the JAX implementation from MeanFlowQL~\Citep{wang2025onestepgenerativepoliciesqlearning}. However, we modified the embedding strategy to assign a unique embedding to each action rather than a single embedding for the entire action chunk, thereby better leveraging the structural capabilities of DiT.

\subsection{Training hyperparameter}

We report the common training parameters for both LPS and the baselines, along with the baseline-specific parameters for OGBench, in \Cref{tab:hy_og}, \Cref{tab:hy_alpha}, and \Cref{tab:hy_w}. We extensively tuned $\alpha$ over the set $\{0.01, 0.03, 0.1, 0.3, 1.0, 3.0, 10.0, 30.0, 100.0, 300.0\}$. Note that the parameter $\alpha$ was tuned individually for each model and latent structure configuration for ablation, except \texttt{visual-task} as we did not include in ablation on latent structure. Additionally, we provide the common training parameters used for the real-world experiments in \Cref{tab:hy_real}.

\begin{table}[ht]
    \centering
    \caption{\footnotesize \textbf{Common parameters for OGBench experiments.}}
    \label{tab:hy_og}
    \begin{tabular}{@{}cc@{}}
        \toprule
        \textbf{Parameter} & \textbf{Value} \\
        \midrule
        Batch size ($M$) &  $256$ \\
        Discount factor ($\gamma$) & $0.99$ \\
        Optimizer & Adam \\
        Learning rate & $3 \times 10^{-4}$ \\
        Learning rate scheduler & constant \\
        Target network update rate ($\tau$) & $5 \times 10^{-3}$ \\
        Critic ensemble size ($K$) & 2 \\
        UTD Ratio & $1$ \\
        Number of flow steps ($T$) & $10$ (Flow matching), $1$ (MeanFlow)\\
        Number of training steps & $10^6$ \\
        Actor network & MLP \\
        Actor network width & $512$ \\
        Actor network depth & $4$ \\
        Critic network & MLP \\
        Critic network width & $256$ \\
        Critic network depth & $4$ \\
        Latent actor network & MLP \\
        Latent actor network width & $256$ \\
        Latent actor network depth & $2$ \\
        Image encoder (\texttt{visual-task}) & \textit{impala small} \\
    \bottomrule
    \end{tabular}
\end{table}

\begin{table}[ht]
    \centering
    \caption{\footnotesize \textbf{Behavior regularization coefficient ($\alpha$).} }
    \label{tab:hy_alpha}
    \begin{tabular}{@{}ccccc@{}}
    \toprule
            Environments (\texttt{normal} / \texttt{sphere})  & {QC-FQL} & {QC-MFQL}  \\
    \midrule
        \texttt{cube-single-*}         & $30.0$ / $10.0$ & $30.0$ / $10.0$  \\
        \texttt{cube-double-*}         & $3.0$ / $3.0$ & $3.0$ / $3.0$  \\
        \texttt{scene-sparse-*}               & $1.0$ / $3.0$ & $1.0$ / $1.0$ \\
        \texttt{puzzle-3x3-sparse-*}          & $3.0$ / $3.0$ & $3.0$ / $3.0$  \\
        \texttt{puzzle-4x4-*}          & $10.0$ / $3.0$ & $3.0$ / $3.0$  \\    
        \texttt{visual-cube-single-task1}          & $10.0$ / - & $30.0$ / -  \\    
        \texttt{visual-cube-double-task1}          & $1.0$ / - & $3.0$ / -  \\    
        \texttt{visual-scene-sparse-task1}          & $30.0$ / - & $3.0$ / -  \\    
        \texttt{visual-puzzle-3x3-sparse-task1}          & $0.1$ / - & $1.0$/ -   \\    
        \texttt{visual-puzzle-4x4-task1}          & $1.0$ / - & $1.0$ / -  \\   
    \bottomrule
    \end{tabular}
\end{table}

\begin{table}[h]
    \centering
    \caption{\footnotesize \textbf{CFG strength ($w$).} }
    \label{tab:hy_w}
    \begin{tabular}{@{}ccccc@{}}
    \toprule
            Environments & {CFGRL} \\
    \midrule    
        \texttt{cube-single-*}         & $1.25$  \\
        \texttt{cube-double-*}         & $2.00$  \\
        \texttt{scene-sparse-*}               & $3.00$ \\
        \texttt{puzzle-3x3-sparse-*}          & $1.50$  \\
        \texttt{puzzle-4x4-*}          & $1.25$  \\    
        \texttt{visual-cube-single-task1}          & $1.25$ \\    
        \texttt{visual-cube-double-task1}          & $2.00$ \\    
        \texttt{visual-scene-sparse-task1}          & $3.00$ \\    
        \texttt{visual-puzzle-3x3-sparse-task1}          & $1.50$ \\    
        \texttt{visual-puzzle-4x4-task1}          & $1.25$ \\    
    \bottomrule
    \end{tabular}
\end{table}

\makeatletter
\setlength{\@fptop}{0pt}
\makeatother

\begin{table}[h]
    \centering
    \caption{\footnotesize \textbf{Common hyperparameters for Real-world experiments.}}
    \label{tab:hy_real}
    \begin{tabular}{@{}cc@{}}
        \toprule
        \textbf{Parameter} & \textbf{Value} \\
        \midrule
        Batch size ($M$) &  $256$ \\
        Discount factor ($\gamma$) & $0.99$ \\
        Optimizer & Adam \\
        Learning rate & $3 \times 10^{-4}$ \\
        Learning rate scheduler & cosine \\
        Target network update rate ($\tau$) & $5 \times 10^{-3}$ \\
        Critic ensemble size ($K$) & 2 \\
        UTD Ratio & $1$ \\
        Number of flow steps ($T$) & $10$ (Flow matching), $1$ (MeanFlow)\\
        Number of training steps & $10^4$ \\
        Actor network & DiT \\
        Actor network hidden dim & $384$ \\
        Actor network depth & $12$ \\
        Actor num heads & $6$ \\
        Critic network & MLP \\
        Critic network width & $512$ \\
        Critic network depth & $4$ \\
        Latent actor network & MLP \\
        Latent actor network width & $512$ \\
        Latent actor network depth & $2$ \\
        Image encoder & \textit{impala} \\
    \bottomrule
    \end{tabular}
\end{table}

\end{document}